\PassOptionsToPackage{table}{xcolor}
\documentclass[11pt]{article}

\usepackage[final]{acl}
\newcommand{\mname}{\textsc{Intralign}}
\newcommand{\fname}{\textsc{Rael}}
\usepackage[table]{xcolor}
\usepackage{times}
\usepackage{amssymb}
\usepackage{latexsym}
\usepackage{booktabs}
\usepackage{mdframed}
\usepackage{fvextra}
\usepackage{float}
\usepackage{colortbl}  
\usepackage{booktabs}
\usepackage{multirow}
\usepackage{amsmath} 

\definecolor{inblue}{RGB}{0, 102, 204}
\definecolor{exgold}{RGB}{204, 153, 0}
\definecolor{confgreen}{RGB}{34, 139, 34}
\usepackage[T1]{fontenc}

\usepackage[utf8]{inputenc}

\usepackage{microtype}

\usepackage{inconsolata}

\usepackage{graphicx}

\title{Transparentize the Internal and External Knowledge Utilization in LLMs with Trustworthy Citation}

\author{Jiajun Shen$^{1,3}$\textsuperscript{\textbf{*}}, 
Tong Zhou$^1$\textsuperscript{\textbf{*}}, 
Yubo Chen$^{1,2}{}^\dag$, 
Delai Qiu$^4$,
Shengping Liu$^4$,
Kang Liu$^{1,2,5}$,
Jun Zhao$^{1,2}$\\
$^1$The Key Laboratory of Cognition and Decision Intelligence for Complex Systems \\
Institute of Automation, Chinese Academy of Sciences\\
$^2$School of Artificial Intelligence, University of Chinese Academy of Sciences\\
$^3$University of Chinese Academy of Sciences\\
$^4$Unisound Al Technology Co,Ltd    $\quad^5$Shanghai Artificial Intelligence Laboratory\\
shenjiajun21@mails.ucas.ac.cn,
tong.zhou@ia.ac.cn\\
\{yubo.chen, kliu, jzhao\}@nlpr.ia.ac.cn, \{qiudelai, liushengping\}@unisound.com
}

\begin{document}
\maketitle

\let\thefootnote\relax\footnotetext{\textsuperscript{\textbf{*}}These authors contributed equally to this work.}
\let\thefootnote\relax\footnotetext{\textsuperscript{\textbf{†}}Corresponding author.}

\begin{abstract}
While hallucinations of large language models could been alleviated through retrieval-augmented generation and citation generation, how the model utilizes internal knowledge is still opaque, and the trustworthiness of its generated answers remains questionable. In this work, we introduce \textcolor{exgold}{Context}-\textcolor{inblue}{Prior} Augmented Citation Generation task, requiring models to generate citations considering both external and internal knowledge while providing trustworthy references, with 5 evaluation metrics focusing on 3 aspects: answer helpfulness, citation faithfulness, and trustworthiness. We introduce \fname{}, the paradigm for our task, and also design \mname, an integrated method containing customary data generation and an alignment algorithm. Our experimental results show that our method achieves a better cross-scenario performance with regard to other baselines. Our extended experiments further reveal that retrieval quality, question types, and model knowledge have considerable influence on the trustworthiness in citation generation.
\end{abstract}
\section{Introduction}
\begin{figure}[t]
  \includegraphics[width=\columnwidth]{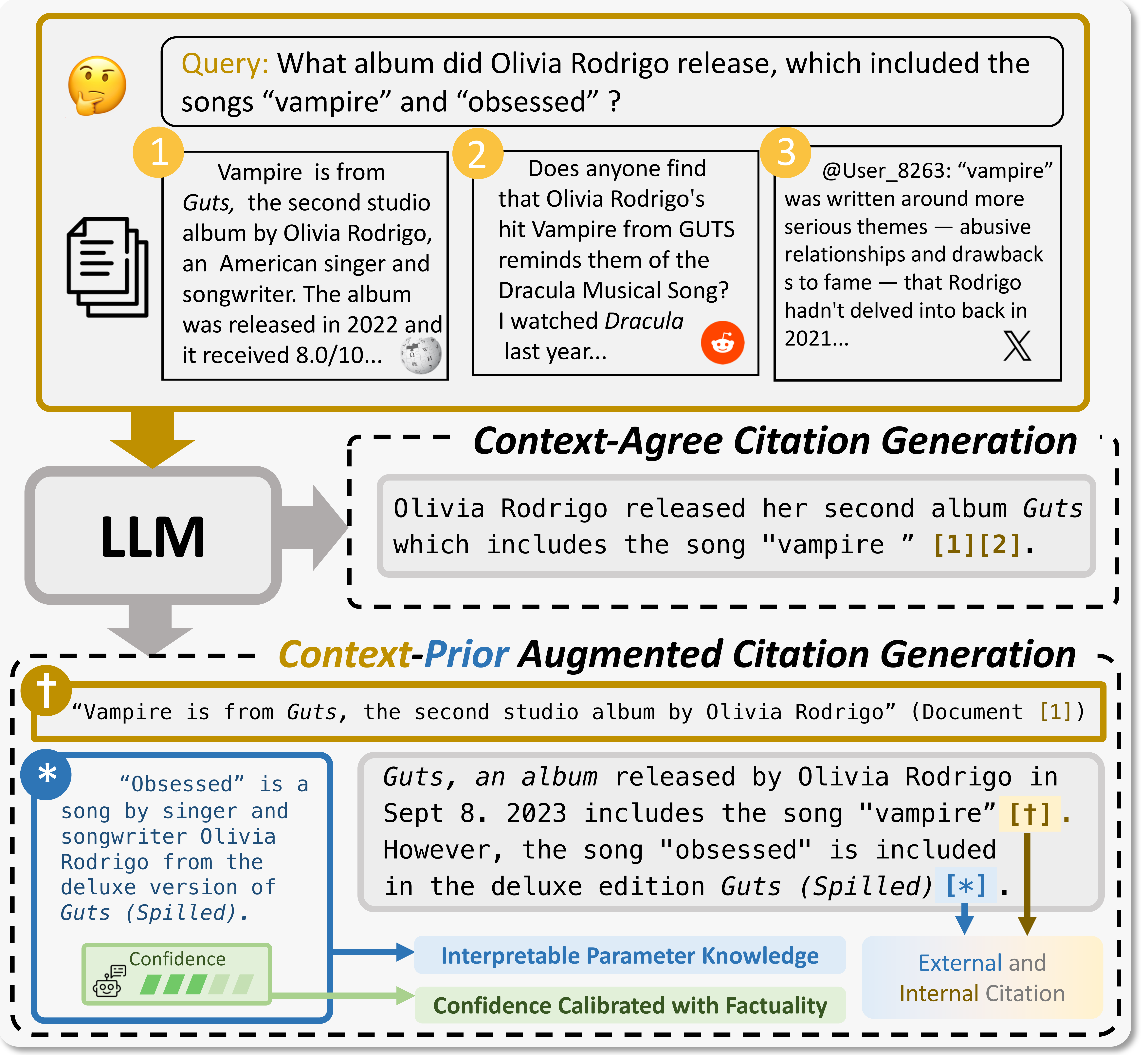}
  \caption{Compared with Context-Agree Citation Generation, the \textcolor{exgold}{Context}-\textcolor{inblue}{Prior} Augmented Citation Generation allows LLMs to appropriately utilize and cite parameter knowledge in an interpretable way, and requires LLMs to extract convincing and concise external references, aiming at transparentize the internal and external knowledge utilization as well as enhancing trustworthiness.}
  \label{fig:tsk}
\end{figure}
Large Language Models (LLMs; \citealp{NEURIPS2020_1457c0d6}) have demonstrated remarkable question-answering (QA) capabilities, providing users with helpful response \citep{shaier-etal-2024-adaptive}. However, due to the hallucination of LLMs, it is crucial to improve the trustworthiness of the responses \citep{liu-etal-2023-evaluating, zhou2024trustworthinessretrievalaugmentedgenerationsystems}. Retrieval Augmented Generation (RAG) with explicit citations can utilize the retrieved external knowledge and link the response to the knowledge to improve the transparency of LLM's response and increase user trust \citep{ding2025citationstrustllmgenerated}. However, previous work (Appendix \ref{rw}) on leveraging context and prior knowledge in LLM generation pays minor attention to the following two problems: (1) the interpretability of LLMs when utilizing prior knowledge and (2) the trustworthiness of the reference.

\textbf{Interpretability of prior knowledge utilization.} Prior knowledge, or parameter knowledge, referring to the knowledge encoded in the model's parameters, could serve as a supplement when the retriever fails to meet the needs of the question in RAG \citep{sun2023recitationaugmented}. For example, as indicated in Figure \ref{fig:tsk}, the question \textit{"What album did Olivia Rodrigo release, which included the songs 'vampire' and 'obsessed'?"} contains two constraints, but the external documents only provide clues about one of them. If the model has supplementary prior knowledge, it can articulate the knowledge, thereby making the answer accurate and providing verifiable evidence. Despite the significance of LLM's prior knowledge, previous studies \citep{yu-etal-2024-revealing, sun2023recitationaugmented, minder2024controllablecontextsensitivityknob, ming2024faithevallanguagemodelstay, cheng2024understandinginterplayparametriccontextual} have largely overlooked its interpretable utilization in citation generation tasks. Therefore, appropriately articulating and citing reliable parameter knowledge in an interpretable way remains challenging.

\textbf{Trustworthiness in reference.} Citations are important in enhancing the convincingness and verifiability of AI-generated content. \citep{ding2025citationstrustllmgenerated}. Highly convincing cited references should contain complete and self-consistent information to support the answer. Concise references reduce the user's verification cost, increasing their trust in the system. For example, Figure \ref{fig:credredun} shows two contrastive cases. The reference \verb|[1]| is concise but lacks background information and supporting evidence, making it doubtful, though easy to verify. The reference \verb|[6]| provides background and details, making it convincing, but includes abundant distracting background information about the question, such as the person's personality, making it less pithy and increasing the cost of verification. The trustworthiness impacts a user's acceptance and trust in the reference, but previous work has not considered this. Furthermore, since there are constraints between convincingness and conciseness, improving both aspects is challenging.

To fill the gap in the interpretability of prior knowledge in citation tasks, we propose \textcolor{exgold}{Context}-\textcolor{inblue}{Prior} Augmented Citation Generation task, which requires the model to generate and cite references from prior knowledge if needed to improve the quality of the response and report a confidence score to transparentize the utilization of prior knowledge. To conduct comprehensive evaluations, we design 5 metrics: (1) Accuracy for the helpfulness of the answer, (2) Citation Recall for citation faithfulness, (3) Convincingness, (4) Conciseness, and (5) Expected Calibration Error for the trustworthiness of the reference. Our evaluations demonstrate a strong correlation between automatic metrics and human judgments.

In response to the requirement of our task, we propose \fname{} (\textbf{R}ational \textbf{A}ttribution and \textbf{El}aboration), a paradigm to enable LLMs to use internal knowledge and generate trustworthy citations appropriately. For the interpretability of parameter knowledge and the faithfulness of citations, we design \mname{} (\textbf{In}terpretable \textbf{Tr}ustworthiness \textbf{Align}ment) to obtain dataset by incorporating the parameter knowledge of the LLM and an alignment step, which enables the LLM to generate faithful and trustworthy citations.

We conduct experiments with different LLM citation generation methods. Our experiments successfully reveal that existing citation generation methods struggle to adapt to scenarios with poor retrieval performance and to cite trustworthy references. \mname{} leads to considerable improvements across all metrics, demonstrating strong practicality. Our contributions can be summarized as follows:

\begin{itemize}
\setlength{\itemsep}{0pt} 
\vspace{-3mm}

\item We propose \textcolor{exgold}{Context}-\textcolor{inblue}{Prior} Augmented Citation Generation task requiring models to appropriately generate and cite references from prior knowledge and design 5 complementary metrics to evaluate helpfulness, faithfulness, and trustworthiness of LLM's response.

\item We introduce \fname{} as the paradigm for our task and \mname{}, which contains multi-scenario trustworthy data generation and interpretability-focused alignment, allowing the model to utilize prior knowledge and generate trustworthy responses.

\item We evaluate 6 baselines and our proposed method with 3 LLMs across 4 scenarios. Experiments reveal the shortcomings of existing methods in improving the overall performance on our task, and our method enables the model to cite references from parameter knowledge while effectively improving the quality of references and enhancing their trustworthiness.
\end{itemize}


\section{Task and Metrics}

In this section, we present a formal definition of the general citation generation task and give the definition of \textcolor{exgold}{Context}-\textcolor{inblue}{Prior} Augmented Citation Generation task along with the metrics introduced.

\textbf{Context-Agree Citation Generation.} The citation generation task accepts a question along with context sequence $D$ and returns an answer, which can be split into $t$ segments $ (S_1, S_2, \dots, S_t)$. Each segment $S_i$ is paired with a reference $R_i $, and we define $ R_i = \mathcal{F}(S_i)$ as segment $S_i$ cites reference $R_i$, or $S_i$ has no citation if $R_i = \varnothing$. Segment $S_i$ is usually split by sentence boundaries \citep{gao-etal-2023-enabling, huang-etal-2024-learning}, and the paired reference is a full document in coarse-grained citations. In fine-grained citation generation \citep{xu2024aliiceevaluatingpositionalfinegrained, zhang-etal-2024-towards-fine-grained}, the reference can span fewer words. To ensure the citation's faithfulness to the context, the cited text must be verbatim: $R_i$ should be a subsequence from $D$. When a question is unanswerable based on $D$, LLM should generate a refusal answer to stay faithfulness to contexts.
 
\textbf{\textcolor{exgold}{Context}-\textcolor{inblue}{Prior} Augmented Citation Generation.} Though faithfulness to the context reduces hallucination, parameter knowledge of LLMs can be beneficial in circumstances of insufficient external sources or low-quality retrieval. To enable LLM to cite parameter knowledge, we require the reference to be from the context or prior knowledge. In our definition, reference $\mathcal{F}(S_i)$ should be either (1) a non-empty extraction from $D_i$ as external reference $\textcolor{exgold}{R^{ex}_i} \neq \varnothing$, or (2) a sequence $(\textcolor{inblue}{R^{in}_i}; \textcolor{confgreen}{P_i})$ as internal reference, where ($;$) denotes concatenation and \textcolor{confgreen}{$P_i$} denotes a confidence score of $\textcolor{inblue}{R^{in}_i} \neq \varnothing$. The confidence score represents the estimation of the factuality of the reference. A refusal answer is preferred if and only if the question is unanswerable based on $D$ and the LLM's parameter knowledge.

\subsection{Metrics}
To implement a comprehensive evaluation in \textcolor{exgold}{Context}-\textcolor{inblue}{Prior} Augmented Citation Generation, we measure \textbf{Accuracy} for the helpfulness of the answer, Citation \textbf{Recall} for the faithfulness of citations, Reference \textbf{Convincingness}, \textbf{Conciseness}, and \textbf{Expected Calibration Error} for the trustworthiness of the reference. We will introduce these metrics below and explain how the metrics ensure robustness against shortcuts in Appendix \ref{app:robust}.

\begin{figure}[t]
  \includegraphics[width=\columnwidth]{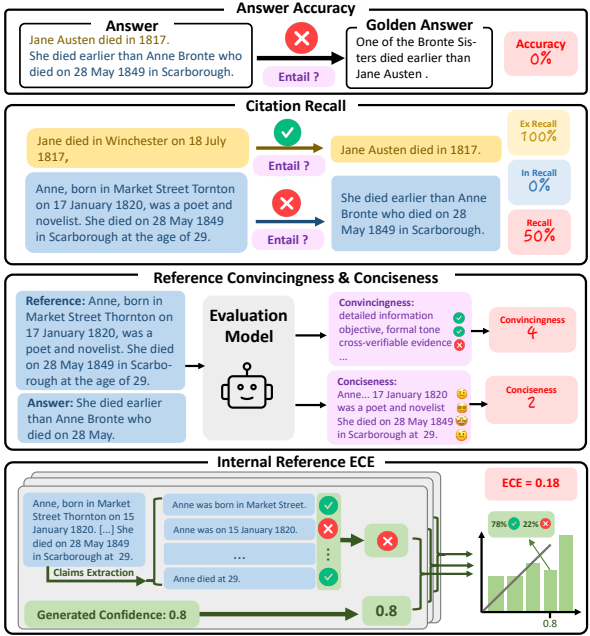}
  \caption{Illustration of our metrics and the auto evaluation process. We use the same NLI model to check entailment to prevent bias.}
  \label{fig:experiments}
\end{figure}

\subsubsection{Answer Accuracy}
Accuracy measures how the response generated by LLM correctly answers the input question. We use a semantics-based method, using an NLI model to verify whether the model's response entails the golden sentence. The NLI model returns a bool value $\phi(a, g)= 1$ if the answer $a$ entails the golden answer $g$. The average accuracy on the dataset is calculated as $\frac{1}{N} \sum_{i = 1}^N  \phi(a_i, g_i)$, where $N$ is the size of the dataset and $a_i, g_i$ are the answer and golden answer from $i$-th sample. We illustrate that our method reduces FN and FP by 25.14\% and 44.93\%, compared to String Exact Match \citep{stelmakh-etal-2022-asqa} in \S \ref{human}, respectively.

\subsubsection{Citation Recall}
Recall shows how faithful the response is to the original reference. Following \citeposs{gao-etal-2023-enabling} work, we also use an NLI model to verify whether the cited reference entails the response. We calculate $Rc^O$ as the overall Recall. Since we observe different recall scores in \textcolor{exgold}{external citations} and \textcolor{inblue}{internal citations}, we also divide $S_i$ where $\mathcal{F}(S_i) \neq \varnothing$ into two sets $\textcolor{exgold}{S^{ex}}$ and $\textcolor{inblue}{S^{in}}$ according to the cited reference, and calculate each type independently as $\textcolor{inblue}{Rc^{in}}$, $\textcolor{exgold}{Rc^{ex}}$. We exclude refusal answers when calculating recall scores. Formally, given a citation function $\mathcal{F}$ and statements $S$, the average Recall scores $Rc^O$ is $Rc^{O} = \frac{1}{|S|} \sum_{S_i \in S} \phi( \mathcal{F}(S_i), S_i)$.

We only use $\textcolor{exgold}{S^{ex}_i} \in \textcolor{exgold}{S^{ex}}$ and $\textcolor{inblue}{S^{in}_i} \in \textcolor{inblue}{S^{in}}$ to calculate $\textcolor{inblue}{Rc^{in}}$ and $\textcolor{exgold}{Rc^{ex}}$, respectively. Sentences without citations ($\mathcal{F}(S_i) = \varnothing)$ is not included in the computation of $\textcolor{inblue}{Rc^{in}}$ and $\textcolor{exgold}{\textcolor{exgold}{Rc^{ex}}}$ but will lower $Rc^O$.

\subsubsection{Reference Convincingness}
Convincingness measures how the cited reference is trustworthy to humans. We expect LLMs to cite convincing references with formal and objective language style, complete expressions, unambiguous entity references, and coherent logic. Fully relying on humans to evaluate this subjective metric is time-consuming, so we use a strong LLM aligned with human preference as the evaluation model for automatic evaluation. When evaluating, we mask all the entities in the reference to avoid bias from prior knowledge and then ask the evaluation model to generate a score from 1 to 5 with an explanation considering the following aspects: tone, style, objectivity, logical coherence, disambiguation, and richness of evidence. We show our prompt in Appendix \ref{prompts}. We also conduct experiments in \S \ref{human} to verify human and automatic evaluation alignment.

\subsubsection{Reference Conciseness}
\begin{figure}[H]
  \includegraphics[width=\columnwidth]{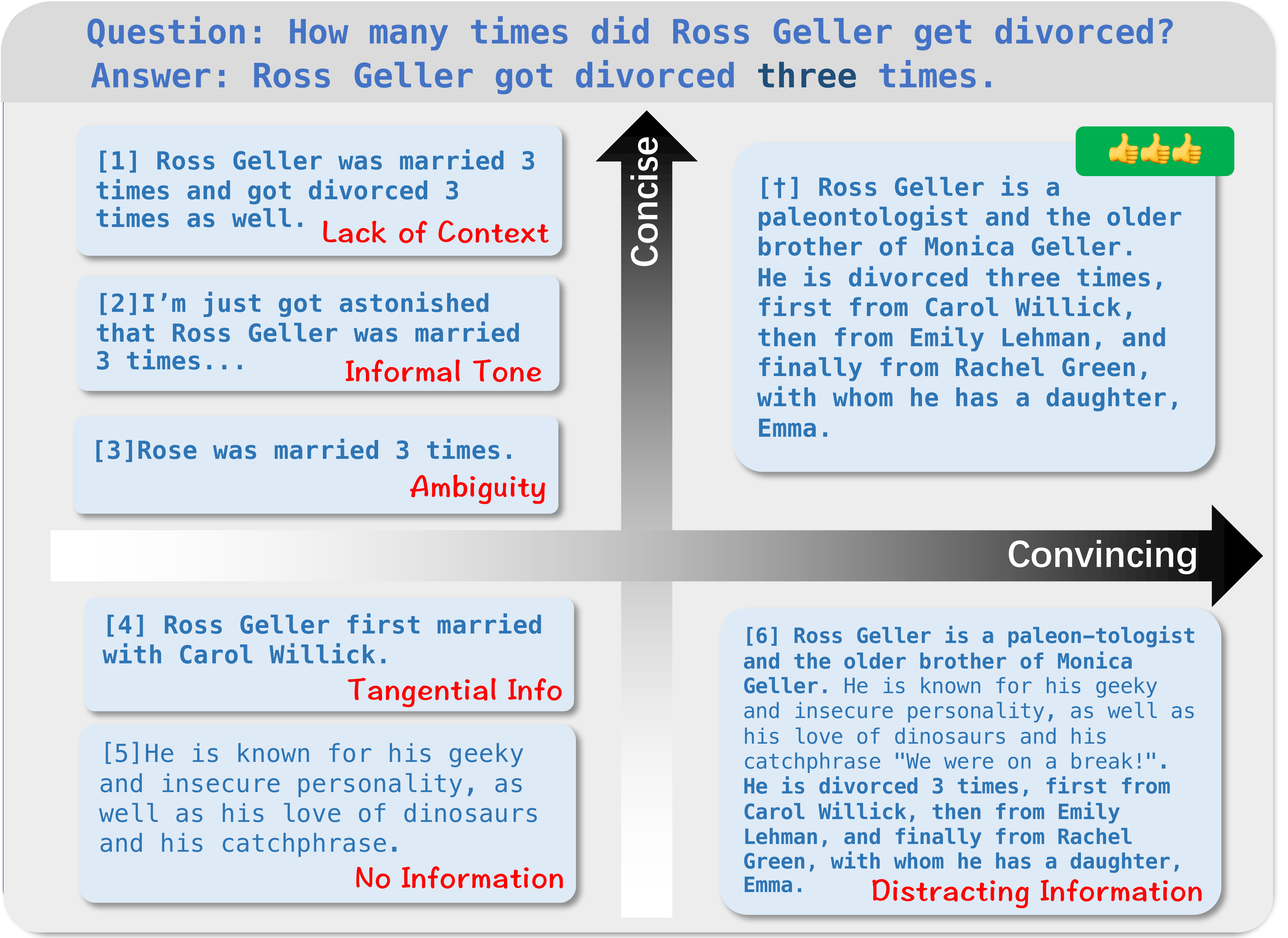}
  \caption{Example of different Convincingness and Conciseness Scores}
  \label{fig:credredun}
\end{figure}

Conciseness reflects the subjective cost required by a person when verifying information. Excessive distracting content, such as too much background information, can lead to wasted time in reading and verification. Conciseness is not the average relevance of sentences to the answer because appropriate background information is helpful. We use an evaluation model to simulate the human process of reading sentence-by-sentence, assessing in sequence whether the text provides useful information with minimal distractions, and report a score from 1 to 5. We show our prompt in Appendix \ref{prompts}. The trade-off between Convincingness and Conciseness, as shown in Figure \ref{fig:credredun}, makes it challenging to improve both metrics simultaneously, even if the reference is already objective and coherent.

\subsubsection{Internal Reference ECE}
In our task, we ask LLMs to report a confidence score when generating a reference from parameter knowledge. To measure the faithfulness of the confidence score generation, we use Expected Calibration Error (ECE) to measure the alignment between the confidence of the output reference \textcolor{confgreen}{$P$} and its real factuality. We evaluate the correctness of each internal reference $\textcolor{inblue}{R_{i}^{in}}$ using \textsc{FactScore} \citep{min2023factscorefinegrainedatomicevaluation} and assume the reference is correct if all the facts are correct (i.e., $\textsc{Fs}(\textcolor{inblue}{R_{i}^{in}}) = 1$, $\textsc{Fs}$ returns the factuality of a reference). We assign each \( i \) in the index set to \( m \) bins, \( B_1, B_2, \dots, B_m \), based on \( \textcolor{confgreen}{P_i} \), where for any \( j \in B_k \), \( \textcolor{confgreen}{P_j} \in \left(\frac{k-1}{m}, \frac{k}{m}\right] \), and then the ECE is calculated as:

\vspace{-5mm}
\[
\text{ECE} = \sum_{m=1}^M \frac{|B_m|}{N} \cdot \left| \text{fact}(B_m) - P(B_m) \right|
\]where $\text{fact}(B_m) = \frac{1}{|B_m|} \sum_{i \in B_m} \mathbb{I}(\textsc{Fs}(\textcolor{inblue}{R_{i}^{in}}) = 1)$ and $P(B_m) = \frac{1}{|B_m|} \sum_{i \in B_m} \textcolor{confgreen}{P_i}$.

\DefineVerbatimEnvironment{CustomVerbatim}{Verbatim}{
  commandchars=\\\{\},
  fontsize=\small,
  formatcom=\color{black},
  breaklines=true
}

\section{Dataset}
We construct the dataset from the three latest RAG datasets: CRAG\citep{yang2024crag}, FRAMES\citep{krishna2024factfetchreasonunified}, and SituatedFaithfulnessEval\citep{huang2024enhancinglargelanguagemodels} (SFE), as they provide diverse, challenging questions. 

We combine the three datasets, equip each data point with 5 retrieved documents, and annotate whether the document is ground truth. (i.e., contains the answer). According to whether the data point contains a \textbf{G}round \textbf{T}ruth document or not, we split the dataset into 2 settings \textbf{GT} and $\overline{\textbf{GT}}$, respectively. Detailed dataset profile and annotation step are shown in Appendix \ref{app:ds}.

\section{Method}
\begin{figure*}[!thp]
  \includegraphics[width=1.0\linewidth]{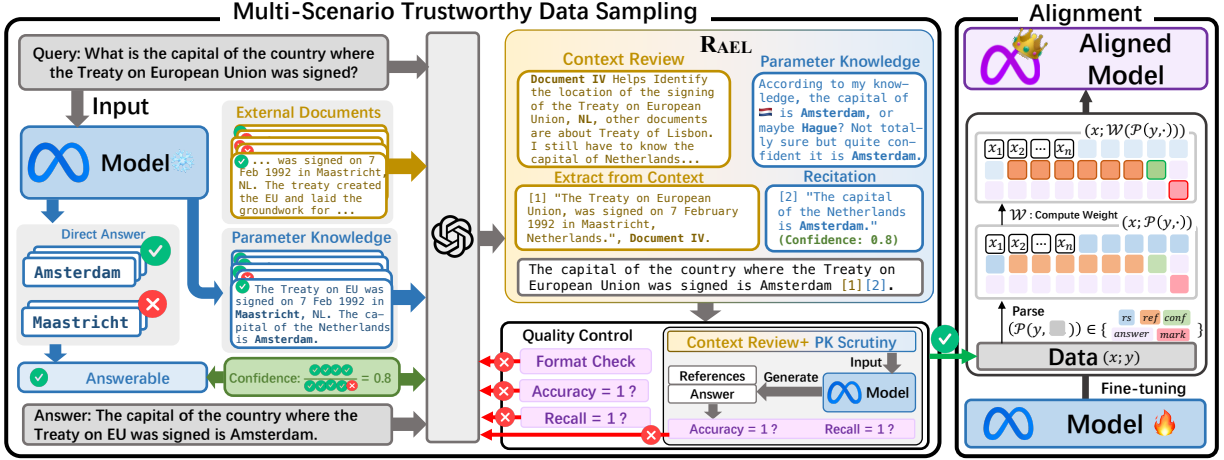}
  \caption {Overview of \mname{}. We first conduct multi-scenario trustworthy data sampling to incorporate parameter knowledge and generate a golden response following our \fname{} paradigm. The verified high-quality data will be used for subsequent Interpretability-Focused Alignment, ultimately resulting in a model capable of utilizing parameter knowledge and generating trustworthy citations.}
  \label{fig:method}
\end{figure*}

Based on the proposed task requirements, we designed a generation paradigm and an integrated method for aligning open-source models. We propose a \textbf{Rational Attribution and Elaboration} (\fname{}) paradigm to align models with the requirements of our proposed task. Specifically, we asked to review the context and scrutinize parameter knowledge to help selectively use context and faithfully state parameter knowledge. Then, the model extracts context and recites parameter knowledge to provide trustworthy references. We design \mname{} (\textbf{In}terpretability-\textbf{Tr}ustworthiness \textbf{Align}ment, a pipeline using reject sampling to generate customary data from GPT-4o \citep{openai2024gpt4technicalreport} and use the data tailored for the specific target model to enhance its performance, as in Figure \ref{fig:method}. 

\subsection{Rational Attribution and Elaboration}
To ensure the model stays faithful to both the internal and external information in generating, we require the model to review the context, scrutinize its own knowledge, and then generate a series of context excerpts or self-generated parameter knowledge as references, along with a final answer containing citations. Reviewing and scrutiny enable the model to rationally consider the sufficiency of external information and the potential relevant knowledge from within, enhancing faithfulness and explainability. The example is shown in Figure \ref{fig:method}, with a Context Review, Parameter Knowledge, References (Including extracted context and recited internal knowledge), and an Answer.





\subsection{Interpretable Trustworthiness Alignment}

To improve the model's faithfulness and trustworthiness for external and internal knowledge in citation generation, we propose a pipeline that samples $k$ documents and $k$ direct answers from the target LLM. If any documents are ground truth or any answer is correct, the LLM is regarded with the necessary parameter knowledge to the question. According to whether the LLM has corresponding \textbf{P}arameter \textbf{K}nowledge, we divide the dataset into two categories \textbf{PK} and $\overline{\textbf{PK}}$, respectively.

For all sampled documents $d_1, d_2,..., d_k$ for each question, we use an NLI model $\phi(\cdot)$ to check whether they contain the answer $g$. The golden confidence is $\frac{\sum_1^k \phi(d_i,g)}{k}$ (The possibility of consistently generating a document with the golden answer). The formula is inspired by self-consistency-based uncertainty measurement \citep{wang2023selfconsistency} in LLMs, but we additionally use the NLI model to ensure factuality.

Given the external documents and internal documents, we asked GPT-4o to generate multiple responses with the \fname{} paradigm using the prompt in Figure \ref{fig:gen prompt}. We evaluate the generated results and regenerate data with incorrect answers or unfaithful citations. Next, we use only the questions, the context review, and parameter knowledge scrutiny to make the target model generate answers and references. We regenerate responses with incorrect answers or unfaithful citations to ensure the quality of the review and scrutiny step. Finally, we rerank the responses by Convincingness and Conciseness and select the one with the highest score.

Due to dispersed optimization objectives in alignment and the neglect caused by limited token usage for citation markers and confidence scores, we adjusted the weights of different types of tokens during the alignment step to make the model focus more on interpretable and trustworthy information. We compute token-wise weighted loss to ensure focus on citation generation. We parse label token sequence in the dataset, and for each token $y_t$ in the label, the parser function $\mathcal{P}$ maps the label sequence $y$ and index $t$ to a type $\tau = \mathcal{P}(y, t) \in \mathcal{T}$, where $\mathcal{T} = \{\tau_{rs}, \tau_{ref}, \tau_{answer}, \tau_{conf}, \tau_{mark}\}$, each denoting review and scrutiny, reference, answer, confidence, and citation markers (such as \verb|[1][2]|). Each type $\tau \in \mathcal{T}$ is assigned with a weight $w = \mathcal{W(\tau)}$. The loss of the \( i \)-th output $ y^{(i)} = (y_1^{(i)}, y_2^{(i)}, \dots, y_{T_i}^{(i)}) $ given the input $x^{(i)}$ is
\vspace{-5mm}

\[
\small
\mathcal{L} = \sum_{t=1}^{T_i} \mathcal{W}(\mathcal{P}(y^{(i)}, t)) \log P_{\theta}(y_t^{(i)} | y_{<t}^{(i)}, x^{(i)})
\]

\vspace{-2mm}

As for the weights, to ensure that the loss gives sufficient attention to confidence, reference and citation markers, we define the total weights for type $\tau$ as $\hat{\mathcal{W}}(\tau) = \sum_{i=0}^{T_{i}} \mathcal{W}(\tau) \cdot\mathbb{I}(\mathcal{P}(y, t) =\tau) $, and set $\hat{\mathcal{W}}(\tau_{conf}) = \hat{\mathcal{W}}(\tau_{ref}), \hat{\mathcal{W}}(\tau_{mark}) = \hat{\mathcal{W}}(\tau_{answer})$. Since our metrics highly depend on the references, we fix the weights of review, scrutiny, and answer to $\mathcal{W}(\tau_{rs})= \mathcal{W}(\tau_{answer})= 1$ and increase the weight of reference such that $\hat{\mathcal{W}}(\tau_{ref}) = \hat{\mathcal{W}}(\tau_{rs}) +  \hat{\mathcal{W}}(\tau_{answer})$ to make the model focus more on reference generation. $\mathcal{W}$ can be determined given the constraints above.

As indicated in Figure \ref{fig:method}, after having determined the weights, we use function $(x;\mathcal{P}(y,\cdot{}))$ to convert the tokenized input into a tensor of labels, and then apply $(x;\mathcal{W}(\mathcal{P}(y,\cdot{}))$ to convert the labels to different weights. We use darkness to represent the weight in the figure (e.g., the orange tokens are darker since they represent the references).

This interpretability-focused alignment with dynamic weights improves the model's trustworthiness and interpretability without sacrificing its overall performance, as shown in the results in \S \ref{ablation}.

\renewcommand{\arraystretch}{0.6} 
\setlength{\abovecaptionskip}{5pt}  

\begin{table*}[h!]
    \centering
    \small
    \resizebox{\textwidth}{!}{%
    \begin{tabular}{p{1.5cm}clccccccc>{\columncolor{myblue}\color{white}}c}
        \toprule
        Scenario &Model & Method & Helpfulness  & \multicolumn{3}{c}{Recall $\uparrow$}  & \multicolumn{3}{c}{Trustworthiness} \\
       \cmidrule(lr){4-4}  \cmidrule(lr){5-7}         \cmidrule(lr){8-10} 
                 & &    & Accuracy $\uparrow$ & $\textcolor{exgold}{Rc^{ex}}$ & $\textcolor{inblue}{Rc^{in}}$ &$Rc^O$   & Conv.  $\uparrow$ & Conc.  $\uparrow$ & \textcolor{confgreen}{ECE}  $\downarrow$   \\

        \cmidrule(lr){2-10} 
       
      \multirow{26}{*}{\fbox{\textbf{GT}, \textbf{PK}}}  &   \multirow{9}{*}{Llama-3.1-8B-Instruct}  
&\textsc{Recite}$^\dagger$&61.38 $_{(2.78)}$&\textcolor{exgold}{-}&\textcolor{inblue}{65.71  $_{(1.02)}$}&60.23  $_{(1.04)}$&3.37 $_{(0.03)}$&3.36 $_{(0.05)}$&\textcolor{confgreen}{0.29 $_{(0.04)}$} \\ 
       
&&\textsc{Front}$^\ddagger$&72.08  $_{(1.49)}$&\textcolor{exgold}{\textbf{71.84}  $_{(1.84)}$}&\textcolor{inblue}{-}&48.00  $_{(1.50)}$&3.42 $_{(0.28)}$&2.01 $_{(0.07)}$&\textcolor{confgreen}{-}\\

&&\textsc{ContextCite}$^\ddagger$&71.10  $_{(4.95)}$&\textcolor{exgold}{25.25  $_{(1.43)}$}&\textcolor{inblue}{ -}&25.08  $_{(1.29)}$&3.43 $_{(0.01)}$&3.47 $_{(0.01)}$&\textcolor{confgreen}{-}\\

&&\textsc{FootNote}&65.91  $_{(2.09)}$&\textcolor{exgold}{51.20  $_{(1.65)}$}&\textcolor{inblue}{44.38 $_{(1.81)}$}&16.03  $_{(1.63)}$&3.32 $_{(0.1)}$&3.96 $_{(0.12)}$&\textcolor{confgreen}{0.20 $_{(0.02)}$}\\

&&\textsc{PostCite}&71.64 $_{(3.88)}$&\textcolor{exgold}{31.01 $_{(3.04)}$}&\textcolor{inblue}{30.77  $_{(2.58)}$}&29.82  $_{(1.84)}$&3.47 $_{(0.08)}$&2.25 $_{(0.12)}$&\textcolor{confgreen}{0.14 $_{(0.02)}$}\\

&&Guided-\fname{}&62.87 $_{(4.43)}$&\textcolor{exgold}{61.54 $_{(2.06)}$}&\textcolor{inblue}{74.46 $_{(1.73)}$}&57.33 $_{(1.49)}$&3.14 $_{(0.09)}$&3.97 $_{(0.05)}$&\textcolor{confgreen}{0.12 $_{(0.03)}$}\\

&&\mname{} (Ours)&\textbf{75.90} $_{(2.80)}$&\textcolor{exgold}{67.63 $_{(1.46)}$}&\textcolor{inblue}{\textbf{85.80} $_{(1.51)}$}&\textbf{63.72} $_{(1.02)}$&\textbf{3.61} $_{(0.11)}$&\textbf{4.05} $_{(0.07)}$&\cellcolor{confgreen!20}\textcolor{confgreen}{\textbf{0.10} $_{(0.02)}$}\\

       \cmidrule(lr){2-10} 
              
       & \multirow{9}{*}{Llama-3.1-70B-Instruct}  &\textsc{Recite}$^\dagger$&45.05 $_{(2.46)}$ & \textcolor{exgold}{-} & \textcolor{inblue}{81.62 $_{(5.15)}$} & 77.86 $_{(3.34)}$ & 3.54 $_{(0.15)}$ & 3.86 $_{(0.06)}$ & \textcolor{confgreen}{0.22 $_{(0.03)}$} \\
       
       &  &\textsc{Front}$^\ddagger$&76.65 $_{(1.53)}$ &\cellcolor{exgold!20} \textcolor{exgold}{\textbf{72.82} $_{(6.34)}$} & \textcolor{inblue}{-} & 57.61 $_{(3.90)}$ & 3.26 $_{(0.07)}$ & 2.53 $_{(0.06)}$ & \textcolor{confgreen}{-} \\
       
      &   &\textsc{ContextCite}$^\ddagger$&72.49 $_{(1.34)}$&\textcolor{exgold}{33.08 $_{(3.92)}$}&\textcolor{inblue}{-}&33.06 $_{(2.23)}$&3.51 $_{(0.03)}$&3.19 $_{(0.09)}$&\textcolor{confgreen}{-}\\
      &   &\textsc{FootNote}&75.83 $_{(4.30)}$ & \textcolor{exgold}{52.91 $_{(3.65)}$} & \textcolor{inblue}{32.14 $_{(1.44)}$} & 30.36 $_{(1.12)}$ & 3.43 $_{(0.14)}$ & 4.10 $_{(0.04)}$ & \textcolor{confgreen}{0.23 $_{(0.00)}$} \\
       &  &\textsc{PostCite}&64.82 $_{(2.09)}$ & \textcolor{exgold}{23.86 $_{(2.32)}$} & \textcolor{inblue}{66.29 $_{(2.13)}$} & 43.71 $_{(1.76)}$ & 3.69 $_{(0.11)}$ & 1.81 $_{(0.10)}$ & \textcolor{confgreen}{0.19 $_{(0.02)}$} \\
      &  &Guided-\fname{}&73.06 $_{(6.20)}$ & \textcolor{exgold}{62.00 $_{(4.93)}$} & \textcolor{inblue}{66.67 $_{(2.11)}$} & 59.79 $_{(1.19)}$ & 3.32 $_{(0.04)}$ & 3.97 $_{(0.03)}$ & \textcolor{confgreen}{0.13 $_{(0.01)}$} \\
      &   &\mname{} (Ours)&\textbf{85.72} $_{(3.13)}$ & \textcolor{exgold}{68.45 $_{(1.20)}$} &\cellcolor{inblue!20} \textcolor{inblue}{\textbf{88.10} $_{(1.01)}$} &\cellcolor{gray!20} \textbf{78.79} $_{(1.27)}$ &\cellcolor{gray!20} \textbf{3.69} $_{(0.19)}$ & \cellcolor{gray!20}\textbf{4.42} $_{(0.14)}$ & \textcolor{confgreen}{\textbf{0.10} $_{(0.01)}$} \\
               
       \cmidrule(lr){2-10} 
       & \multirow{4}{*}{GPT-4o}  
        &\textsc{PostCite}&81.81 $_{(6.18)}$ & \textcolor{exgold}{39.54 $_{(3.17)}$} & \textcolor{inblue}{\textbf{81.95} $_{(2.46)}$} & 56.32 $_{(1.79)}$ & 2.94 $_{(0.14)}$ & 2.95 $_{(0.04)}$ & \textcolor{confgreen}{0.20 $_{(0.02)}$} \\
      &  &\textsc{FootNote}&\textbf{82.19} $_{(5.45)}$&\textcolor{exgold}{57.22 $_{(6.85)}$}&\textcolor{inblue}{52.14 $_{(4.27)}$}&51.75 $_{(5.60)}$&3.42 $_{(0.04)}$&3.61 $_{(0.16)}$&\textcolor{confgreen}{0.18 $_{(0.02)}$}\\
      &  &Guided-\fname{}&81.59 $_{(10.76)}$&\textcolor{exgold}{\textbf{62.94} $_{(7.31)}$}&\textcolor{inblue}{66.67 $_{(6.62)}$}&\textbf{58.84} $_{(7.01)}$&\textbf{3.58} $_{(0.07)}$&\textbf{4.00} $_{(0.14)}$&\textcolor{confgreen}{0.10 $_{(0.00)}$}\\
    
       \cmidrule(lr){2-10} 
       & \multirow{2}{*}{DeepSeek-R1}  
      &\textsc{FootNote}&\cellcolor{gray!20}90.55 $_{(-)}$&\textcolor{exgold}{56.78 $_{(-)}$}&\textcolor{inblue}{50.33 $_{(-)}$}&51.25 $_{(-)}$&3.20 $_{(-)}$&3.87 $_{(-)}$&\textcolor{confgreen}{0.17 $_{(-)}$}\\
      & &\textsc{Guided-RAEL}&84.77 $_{(-)}$&\textcolor{exgold}{53.89 $_{(-)}$}&\textcolor{inblue}{49.47 $_{(-)}$}&51.45 $_{(-)}$&3.28 $_{(-)}$&4.04 $_{(-)}$&\textcolor{confgreen}{0.15 $_{(-)}$}\\

       \cmidrule(lr){2-10} 
       & \multirow{2}{*}{o1-mini}  
      &\textsc{Footnote}&77.53 $_{(-)}$&\textcolor{exgold}{53.92 $_{(-)}$}&\textcolor{inblue}{43.56 $_{(-)}$}&49.01 $_{(-)}$&3.26 $_{(-)}$&3.89 $_{(-)}$&\textcolor{confgreen}{0.15 $_{(-)}$}\\
&             &\textsc{Guided-RAEL}&70.88 $_{(-)}$&\textcolor{exgold}{56.25 $_{(-)}$}&\textcolor{inblue}{47.92 $_{(-)}$}&52.17 $_{(-)}$&3.23 $_{(-)}$&4.00 $_{(-)}$&\textcolor{confgreen}{0.13 $_{(-)}$}\\

        \midrule

    \multirow{26}{*}{\fbox{\textbf{GT}, $\overline{\textbf{PK}}$}}&
         \multirow{9}{*}{Llama-3.1-8B-Instruct}&\textsc{Recite}$^\dagger$&0.95 $_{(1.41)}$&\textcolor{exgold}{-}&\textcolor{inblue}{56.51  $_{(1.50)}$}&48.49  $_{(1.88)}$&3.05 $_{(0.03)}$&2.04 $_{(0.03)}$&\textcolor{confgreen}{0.23 $_{(0.04)}$} \\
       
        & &\textsc{Front}$^\ddagger$&64.29  $_{(1.66)}$&\textcolor{exgold}{\textbf{68.62}  $_{(2.82)}$}&\textcolor{inblue}{-}&56.25 $_{(1.88)}$&1.80 $_{(0.05)}$&1.86 $_{(0.05)}$&\textcolor{confgreen}{-}\\
       
         &&\textsc{ContextCite}$^\ddagger$&56.67 $_{(3.49)}$&\textcolor{exgold}{35.56 $_{(1.14)}$}&\textcolor{inblue}{-}&35.55 $_{(1.20)}$&3.51 $_{(0.16)}$&3.01 $_{(0.03)}$&\textcolor{confgreen}{-}\\
         &&\textsc{FootNote}&50.95  $_{(3.81)}$&\textcolor{exgold}{56.52 $_{(1.90)}$}&\textcolor{inblue}{45.90 $_{(1.19)}$}&14.16 $_{(1.23)}$&3.55 $_{(0.02)}$&3.47 $_{(0.07)}$&\textcolor{confgreen}{0.17 $_{(0.02)}$}\\
         &&\textsc{PostCite}&63.33 $_{(2.51)}$&\textcolor{exgold}{22.03 $_{(1.08)}$}&\textcolor{inblue}{35.93 $_{(2.03)}$}&24.24 $_{(1.59)}$&3.60 $_{(0.01)}$&2.38 $_{(0.20)}$&\textcolor{confgreen}{0.13 $_{(0.01)}$}\\
        &&Guided-\fname{}&48.10  $_{(2.45)}$&\textcolor{exgold}{51.61 $_{(1.77)}$}&\textcolor{inblue}{43.75 $_{(1.42)}$}&43.56 $_{(1.39)}$&3.48 $_{(0.10)}$&\textbf{3.80} $_{(0.08)}$&\textcolor{confgreen}{0.13 $_{(0.03)}$}\\
         &&\mname{} (Ours)&\textbf{69.05} $_{(2.65)}$&\textcolor{exgold}{44.53  $_{(1.43)}$}&\textcolor{inblue}{\textbf{87.95} $_{(1.72)}$}&\textbf{51.84} $_{(1.01)}$&\textbf{3.64} $_{(0.03)}$&3.78 $_{(0.12)}$&\textcolor{confgreen}{\textbf{0.11} $_{(0.02)}$}\\

        \cmidrule(lr){2-10} 
               
         &\multirow{9}{*}{Llama-3.1-70B-Instruct}&\textsc{Recite}$^\dagger$&1.95 $_{(0.25)}$ & \textcolor{exgold}{-} & \textcolor{inblue}{80.18 $_{(6.52)}$} & 78.73 $_{(6.17)}$ & \cellcolor{gray!20}\textbf{3.82} $_{(0.1)}$ & 3.09 $_{(0.07)}$ & \textcolor{confgreen}{0.23 $_{(0.02)}$} \\
       
         &&\textsc{Front}$^\ddagger$&65.24 $_{(0.55)}$ &\cellcolor{exgold!20}  \textcolor{exgold}{\textbf{72.10} $_{(0.69)}$} & \textcolor{inblue}{-} & 53.78 $_{(5.35)}$ & 3.48 $_{(0.11)}$ & 2.32 $_{(0.07)}$ & \textcolor{confgreen}{-} \\
       
         &&\textsc{ContextCite}$^\ddagger$&63.07 $_{(3.17)}$&\textcolor{exgold}{38.89 $_{(5.90)}$}&\textcolor{inblue}{-}&38.86 $_{(0.59)}$&3.32 $_{(0.13)}$&3.29 $_{(0.04)}$&\textcolor{confgreen}{-}\\
         &&\textsc{FootNote}&67.74 $_{(6.48)}$ & \textcolor{exgold}{68.18 $_{(1.97)}$} & \textcolor{inblue}{33.33 $_{(3.06)}$} & 30.92 $_{(1.32)}$ & 3.71 $_{(0.12)}$ & 4.40 $_{(0.13)}$ & \textcolor{confgreen}{0.18 $_{(0.01)}$} \\
         &&\textsc{PostCite}&58.13 $_{(1.04)}$ & \textcolor{exgold}{21.93 $_{(0.63)}$} & \textcolor{inblue}{68.90 $_{(6.98)}$} & 42.56 $_{(2.21)}$ & 3.86 $_{(0.10)}$ & 1.37 $_{(0.02)}$ & \textcolor{confgreen}{0.15 $_{(0.01)}$} \\
        &&Guided-\fname{}&63.40 $_{(0.79)}$ & \textcolor{exgold}{40.44 $_{(4.71)}$} & \textcolor{inblue}{50.00 $_{(0.26)}$} & 40.61 $_{(7.31)}$ & 3.54 $_{(0.06)}$ & 3.18 $_{(0.02)}$ & \textcolor{confgreen}{\cellcolor{confgreen!20}\textbf{0.10} $_{(0.00)}$} \\
         &&\mname{} (Ours)&\textbf{71.25} $_{(1.44)}$ & \textcolor{exgold}{54.42 $_{(2.60)}$} & \textcolor{inblue}{\cellcolor{inblue!20}\textbf{82.07} $_{(3.76)}$} & \textbf{\cellcolor{gray!20}60.32} $_{(4.79)}$ & 3.60 $_{(0.08)}$ & \textbf{4.47} $_{(0.03)}$ & \textcolor{confgreen}{0.13 $_{(0.03)}$} \\
                
        \cmidrule(lr){2-10} 
               
        &\multirow{3}{*}{GPT-4o}&\textsc{PostCite}&\textbf{75.00} $_{(2.87)}$ & \textcolor{exgold}{24.44 $_{(1.03)}$} & \textcolor{inblue}{\textbf{80.67} $_{(7.68)}$} & \textbf{52.03} $_{(2.99)}$ & \textbf{3.78} $_{(0.12)}$ & 2.67 $_{(0.08)}$ & \textcolor{confgreen}{0.18 $_{(0.01)}$} \\
        &&\textsc{FootNote}&66.03 $_{(6.02)}$&\textcolor{exgold}{\textbf{47.88} $_{(0.03)}$}&\textcolor{inblue}{52.87 $_{(2.13)}$}&32.38 $_{(2.8)}$&3.55 $_{(0.00)}$&3.85 $_{(0.09)}$&\textcolor{confgreen}{0.20 $_{(0.03)}$}\\
        &&Guided-\fname{}&69.51 $_{(5.2)}$&\textcolor{exgold}{41.00 $_{(3.45)}$}&\textcolor{inblue}{83.33 $_{(1.78)}$}&42.94 $_{(4.43)}$&3.47 $_{(0.11)}$&\textbf{4.09} $_{(0.05)}$&\textcolor{confgreen}{0.18 $_{(0.02)}$}\\

       \cmidrule(lr){2-10} 
       & \multirow{2}{*}{DeepSeek-R1}  
      &\textsc{Footnote}&\cellcolor{gray!20}76.02 $_{(-)}$&\textcolor{exgold}{41.66 $_{(-)}$}&\textcolor{inblue}{40.78 $_{(-)}$}&40.96 $_{(-)}$&3.35 $_{(-)}$&3.82 $_{(-)}$&\textcolor{confgreen}{0.18 $_{(-)}$}\\
      & &\textsc{Guided-RAEL}&76.64 $_{(-)}$&\textcolor{exgold}{43.05 $_{(-)}$}&\textcolor{inblue}{57.14 $_{(-)}$}&41.89 $_{(-)}$&3.26 $_{(-)}$&4.00 $_{(-)}$&\textcolor{confgreen}{0.18 $_{(-)}$}\\

       \cmidrule(lr){2-10} 
       & \multirow{2}{*}{o1-mini}  
      &\textsc{Footnote}&64.61 $_{(-)}$&\textcolor{exgold}{47.61 $_{(-)}$}&\textcolor{inblue}{43.06 $_{(-)}$}&45.51 $_{(-)}$&3.60 $_{(-)}$&4.05 $_{(-)}$&\textcolor{confgreen}{0.15 $_{(-)}$}\\
&             &\textsc{Guided-RAEL}&51.58 $_{(-)}$&\textcolor{exgold}{57.00 $_{(-)}$}&\textcolor{inblue}{40.23 $_{(-)}$}&38.53 $_{(-)}$&3.53 $_{(-)}$&\cellcolor{gray!20}4.51 $_{(-)}$&\textcolor{confgreen}{0.15 $_{(-)}$}\\

        \midrule
               
            \multirow{26}{*}{\fbox{$\overline{\textbf{GT}}$, \textbf{PK}}}&
         \multirow{9}{*}{Llama3.1-8B-Instruct}&\textsc{Recite}$^\dagger$&61.49  $_{(1.68)}$&\textcolor{exgold}{-}&\textcolor{inblue}{66.13  $_{(1.17)}$}&59.81  $_{(2.32)}$&3.32 $_{(0.02)}$&3.32 $_{(0.03)}$&\textcolor{confgreen}{0.30 $_{(0.03)}$} \\
       
         &&\textsc{Front}$^\ddagger$&48.60  $_{(3.07)}$&\textcolor{exgold}{54.61  $_{(1.28)}$}&\textcolor{inblue}{-}&31.60 $_{(1.56)}$&3.30 $_{(0.08)}$&1.86 $_{(0.01)}$&\textcolor{confgreen}{-}\\
        &&\textsc{ContextCite}$^\ddagger$&44.41 $_{(3.99)}$&\textcolor{exgold}{12.46 $_{(1.47)}$}&\textcolor{inblue}{-}&12.46 $_{(1.41)}$&3.37 $_{(0.16)}$&3.81 $_{(0.15)}$&\textcolor{confgreen}{-}\\
         &&\textsc{FootNote}&35.44 $_{(1.30)}$&\textcolor{exgold}{35.46 $_{(1.19)}$}&\textcolor{inblue}{31.15 $_{(2.11)}$}&18.87 $_{(1.33)}$&3.59 $_{(0.11)}$&3.29 $_{(0.08)}$&\textcolor{confgreen}{0.17 $_{(0.01)}$}\\

         &&\textsc{PostCite}&48.60 $_{(3.98)}$&\textcolor{exgold}{24.70 $_{(2.43)}$}&\textcolor{inblue}{33.84 $_{(2.47)}$}&28.59 $_{(2.26)}$&3.62 $_{(0.11)}$&2.51 $_{(0.12)}$&\textcolor{confgreen}{0.20 $_{(0.01)}$}\\
        &&Guided-\fname{}&49.30  $_{(2.97)}$&\textcolor{exgold}{42.77  $_{(3.42)}$}&\textcolor{inblue}{72.34  $_{(2.05)}$}&47.54  $_{(1.82)}$&3.05 $_{(0.15)}$&3.59 $_{(0.10)}$&\textcolor{confgreen}{0.13 $_{(0.03)}$}\\
         &&\mname{} (Ours)&\textbf{62.24}  $_{(3.84)}$&\textcolor{exgold}{\textbf{59.11}  $_{(1.25)}$}&\textcolor{inblue}{\textbf{72.65}  $_{(2.73)}$}&\textbf{59.80}  $_{(1.50)}$&\textbf{3.65} $_{(0.13)}$&\textbf{3.83} $_{(0.07)}$&\textcolor{confgreen}{\textbf{0.11} $_{(0.02)}$}\\
               
        \cmidrule(lr){2-10} 
               
        &\multirow{9}{*}{Llama3.1-70B-Instruct}  &\textsc{Recite}$^\dagger$&44.43 $_{(2.01)}$ & \textcolor{exgold}{-} & \textcolor{inblue}{82.12 $_{(6.18)}$} & 72.86 $_{(5.04)}$ & 3.44 $_{(0.26)}$ & 3.72 $_{(0.19)}$ & \textcolor{confgreen}{0.22 $_{(0.02)}$} \\
       
         &&\textsc{Front}$^\ddagger$&56.06 $_{(0.18)}$ & \textcolor{exgold}{47.46 $_{(0.52)}$} & \textcolor{inblue}{-} & 42.28 $_{(3.96)}$ & 3.32 $_{(0.02)}$ & 2.24 $_{(0.04)}$ & \textcolor{confgreen}{-} \\
        &&\textsc{ContextCite}$^\ddagger$&52.91 $_{(1.65)}$&\textcolor{exgold}{15.81 $_{(2.19)}$}&\textcolor{inblue}{-}&15.81 $_{(1.62)}$&3.55 $_{(0.02)}$&3.86 $_{(0.11)}$&\textcolor{confgreen}{-}\\
         &&\textsc{FootNote}&54.66 $_{(2.08)}$ & \textcolor{exgold}{41.98 $_{(0.58)}$} & \textcolor{inblue}{26.23 $_{(0.77)}$} & 20.62 $_{(0.07)}$ & 3.53 $_{(0.11)}$ & 4.10 $_{(0.06)}$ & \textcolor{confgreen}{0.18 $_{(0.01)}$} \\

         &&\textsc{PostCite}&53.88 $_{(1.64)}$ & \textcolor{exgold}{22.63 $_{(3.17)}$} & \textcolor{inblue}{74.21 $_{(3.48)}$} & 45.13 $_{(1.34)}$ & 3.66 $_{(0.04)}$ & 1.74 $_{(0.05)}$ & \textcolor{confgreen}{0.19 $_{(0.01)}$} \\
        &&Guided-\fname{}&57.32 $_{(7.12)}$ & \textcolor{exgold}{46.15 $_{(1.25)}$} & \textcolor{inblue}{81.94 $_{(7.02)}$} & 51.64 $_{(4.48)}$ & 3.33 $_{(0.09)}$ & 3.18 $_{(0.06)}$ & \textcolor{confgreen}{0.12 $_{(0.00)}$} \\
         &&\mname{} (Ours)&\textbf{75.64} $_{(0.72)}$ & \textcolor{exgold}{\textbf{62.01} $_{(4.20)}$} & \textcolor{inblue}{\cellcolor{inblue!20}\textbf{89.82} $_{(1.35)}$} & \cellcolor{gray!20}\textbf{75.71} $_{(2.19)}$ & \cellcolor{gray!20}\textbf{3.67} $_{(0.07)}$ &\cellcolor{gray!20} \textbf{4.42} $_{(0.09)}$ & \textcolor{confgreen}{\cellcolor{confgreen!20}\textbf{0.09} $_{(0.00)}$} \\
                
        \cmidrule(lr){2-10} 
               
        &\multirow{3}{*}{GPT-4o}&\textsc{PostCite}&\cellcolor{gray!20}\textbf{78.59} $_{(5.84)}$ & \textcolor{exgold}{39.83 $_{(0.81)}$} & \textcolor{inblue}{\textbf{77.26} $_{(1.35)}$} & 53.72 $_{(1.65)}$ & 3.53 $_{(0.13)}$ & 2.94 $_{(0.04)}$ & \textcolor{confgreen}{0.15 $_{(0.00)}$} \\
        &&\textsc{FootNote}&72.89 $_{(6.65)}$&\textcolor{exgold}{40.06 $_{(2.91)}$}&\textcolor{inblue}{53.78 $_{(5.91)}$}&49.51 $_{(8.93)}$&3.42 $_{(0.15)}$&4.00 $_{(0.03)}$&\textcolor{confgreen}{0.15 $_{(0.00)}$}\\
        &&Guided-\fname{}&72.27 $_{(3.13)}$&\textcolor{exgold}{\textbf{43.97} $_{(0.59)}$}&\textcolor{inblue}{69.61 $_{(0.82)}$}&\textbf{54.93} $_{(6.22)}$&\textbf{3.64} $_{(0.08)}$&\textbf{4.04} $_{(0.07)}$&\textcolor{confgreen}{0.22 $_{(0.03)}$}\\

       \cmidrule(lr){2-10} 
       & \multirow{2}{*}{DeepSeek-R1}  
      &\textsc{Footnote}&71.66 $_{(-)}$&\textcolor{exgold}{36.41 $_{(-)}$}&\textcolor{inblue}{50.82 $_{(-)}$}&40.40 $_{(-)}$&3.30 $_{(-)}$&3.86 $_{(-)}$&\textcolor{confgreen}{0.18 $_{(-)}$}\\
&             &\textsc{Guided-RAEL}&73.29 $_{(-)}$&\textcolor{exgold}{36.89 $_{(-)}$}&\textcolor{inblue}{50.32 $_{(-)}$}&42.31 $_{(-)}$&3.20 $_{(-)}$&3.94 $_{(-)}$&\textcolor{confgreen}{0.13 $_{(-)}$}\\

       \cmidrule(lr){2-10} 
       & \multirow{2}{*}{o1-mini}  
      &\textsc{Footnote}&61.23 $_{(-)}$&\textcolor{exgold}{39.82 $_{(-)}$}&\textcolor{inblue}{42.50 $_{(-)}$}&40.83 $_{(-)}$&3.01 $_{(-)}$&3.81 $_{(-)}$&\textcolor{confgreen}{0.13 $_{(-)}$}\\
&             &\textsc{Guided-RAEL}&63.23 $_{(-)}$&\cellcolor{exgold!20}\textcolor{exgold}{69.23 $_{(-)}$}&\textcolor{inblue}{51.02 $_{(-)}$}&45.35 $_{(-)}$&3.09 $_{(-)}$&4.07 $_{(-)}$&\textcolor{confgreen}{0.13 $_{(-)}$}\\

        \bottomrule
    \end{tabular}}
    \setlength{\fboxsep}{1pt}
    \caption{Results on test sets \textbf{GT}, \textbf{PK}; \textbf{GT}, $\overline{\textbf{PK}}$ and $\overline{\textbf{GT}}$, \textbf{PK}. We use different random seeds to run three experiments for each setting (except for o1-like models) and show the mean scores. The values in brackets represent the standard deviation. Methods marked with $\dagger$ are limited to citing parameter knowledge, while sections marked with $\ddagger$ are limited to citing external knowledge. \textbf{bold} values represent model-wise best score, and \colorbox{gray!20}{background} represents the best score (before rounded to 2 decimals) across models. }

    \label{maintable1}
\end{table*}

\setlength{\belowcaptionskip}{0pt}
\section{Experiments}

We conduct comprehensive experiments on three LLMs with baselines and our method. Then, we present the results along with a detailed and thorough analysis.
\vspace{-1mm}
\subsection{Settings}
To showcase the influence of parameter knowledge and instruction following ability, we use two open-source models from the same family with two parameter sizes: Llama-3.1-8B-Instruct and Llama-3.1-70B-Instruct \citep{llama3modelcard}. We also apply a more powerful closed-source LLM, GPT-4o from OpenAI \cite{openai2024gpt4technicalreport}. As deep thinking CoTs in inference like OpenAI-o1 and DeepSeek-R1 models are popular, and their “thinking” part contains the description about how they use external knowledge and internal knowledge, we also use o1-mini and DeepSeek-R1 as our models. We ask the LLM to provide an answer in the RAEL framework, without giving examples for the reasoning step, to make sure the reference model can freely choose their own thinking style.

We obtain 1K training and 0.5K test data for training and evaluation, using GPT-4o-mini to evaluate Convincingness and Conciseness.

\subsection{Baselines}
We compare our method with the state-of-the-art methods on RAG and citation generation. We use 6 baselines in our experiments, including: (1) Guided-\fname{} uses a two-shot prompt to guide the model in applying \fname{} paradigm. (2) \textsc{FootNote} generates answers with reference in the format \verb|\footnote[confidence]{reference}|. (3) \textsc{PostCite} retrieves documents using a GTR retriever and cites the document with the highest similarity score unless it falls below a threshold, in which case the model generates an internal citation. (4) \textsc{Recitation Augmented Generation} samples passages from the model’s parameters and generates answers based on them, determining the final answer via majority voting. (5) \textsc{Front} optimizes citation quality by extracting supporting quotes and ensuring consistency. (6) \textsc{ContextCite} identifies the specific parts of the context that contribute to the response using sparse linear modeling. Details about baselines are shown in \ref{app:baseline}.

\vspace{-3mm}
\subsection{Main Results}
We show our main results on the \textbf{GT}, \textbf{PK}; \textbf{GT}, $\overline{\textbf{PK}}$ and $\overline{\textbf{GT}}$, \textbf{PK} sets in Table \ref{maintable1} and detail the findings below. We also provide further analysis of refusal answers on the $\overline{\textbf{GT}}$, $\overline{\textbf{PK}}$ set in Appendix \ref{abstention}.

\textbf{A notable difference lies between {\textcolor{exgold}{$\mathbf{Rc^{ex}}$}} and \textcolor{inblue}{$\mathbf{Rc^{in}}$}.} For methods utilizing both external and internal citation except \textsc{FootNote}, the internal citation recall \textcolor{inblue}{$Rc^{in}$} is generally higher than the external citation recall \textcolor{exgold}{$Rc^{ex}$}. This indicates that models are more faithful in citation when using internal knowledge, and the \verb|\footnote| format is difficult for LLMs to follow when citing prior knowledge.

\textbf{Existing methods struggle with cross-scenario performance and trustworthiness.} \textsc{Front} and \textsc{ContextCite} suffer from a nearly 30\% drop in accuracy and \textcolor{exgold}{$Rc^{ex}$} when the retrieval quality is low. \textsc{RECITE}, \textsc{Front}, and \textsc{ContextCite} also have shortcomings in terms of trustworthiness, exhibiting lower Convincingness, Conciseness, and higher ECE.

\textbf{Our method achieves better overall performance with
 more trustworthy references.} Our approach achieves outstanding performance across all major metrics in each scenario. Notably, in terms of citation quality, our method minimizes the proportion of uncited sentences as much as possible while maintaining high recall scores for both external and internal citations with \textcolor{inblue}{$Rc^{in}$} higher than 80\%. Additionally, the references generated by our method exhibit better trustworthiness, with a score of nearly or higher than 4.00 in terms of the Conciseness score. The alignment between the model's confidence in parameter knowledge and factual accuracy in our method surpasses most baselines, indicating that the references produced by our model are of higher quality and more trustworthy.

\textbf{LLMs can learn to use external and internal knowledge adaptively.} Our method maintains high accuracy across different scenarios. In the $\textbf{GT}, \overline{\textbf{PK}}$ and $\overline{\textbf{GT}}$, \textbf{PK} sets, the model achieves similarly high accuracy. This demonstrates that our method enables the model to leverage both external and internal knowledge adaptively.

It is noticeable that the inference model achieves relatively higher accuracy but fails to provide high-quality citations, especially for internal citations. We thank the reviewer for reminding us of the o1-like model's ability to give a thoughtful and accurate answer. However, the claim that the general CoT process does not necessarily improve citation quality and trustworthiness still holds. Moreover, given the fact that the R1 model outputs 5 times more tokens than our model (see the table below) due to overthinking, the deep thinking model's performance on our task is still limited.

\subsection{In-depth Results of the Task}

In this section, we discuss more experimental results on different settings that reveal the characteristics of models on the Context-Prior Augmented Citation Generation task.
\subsubsection{Results on a less credible external source}
\begin{figure}[t]
  \includegraphics[width=\columnwidth]{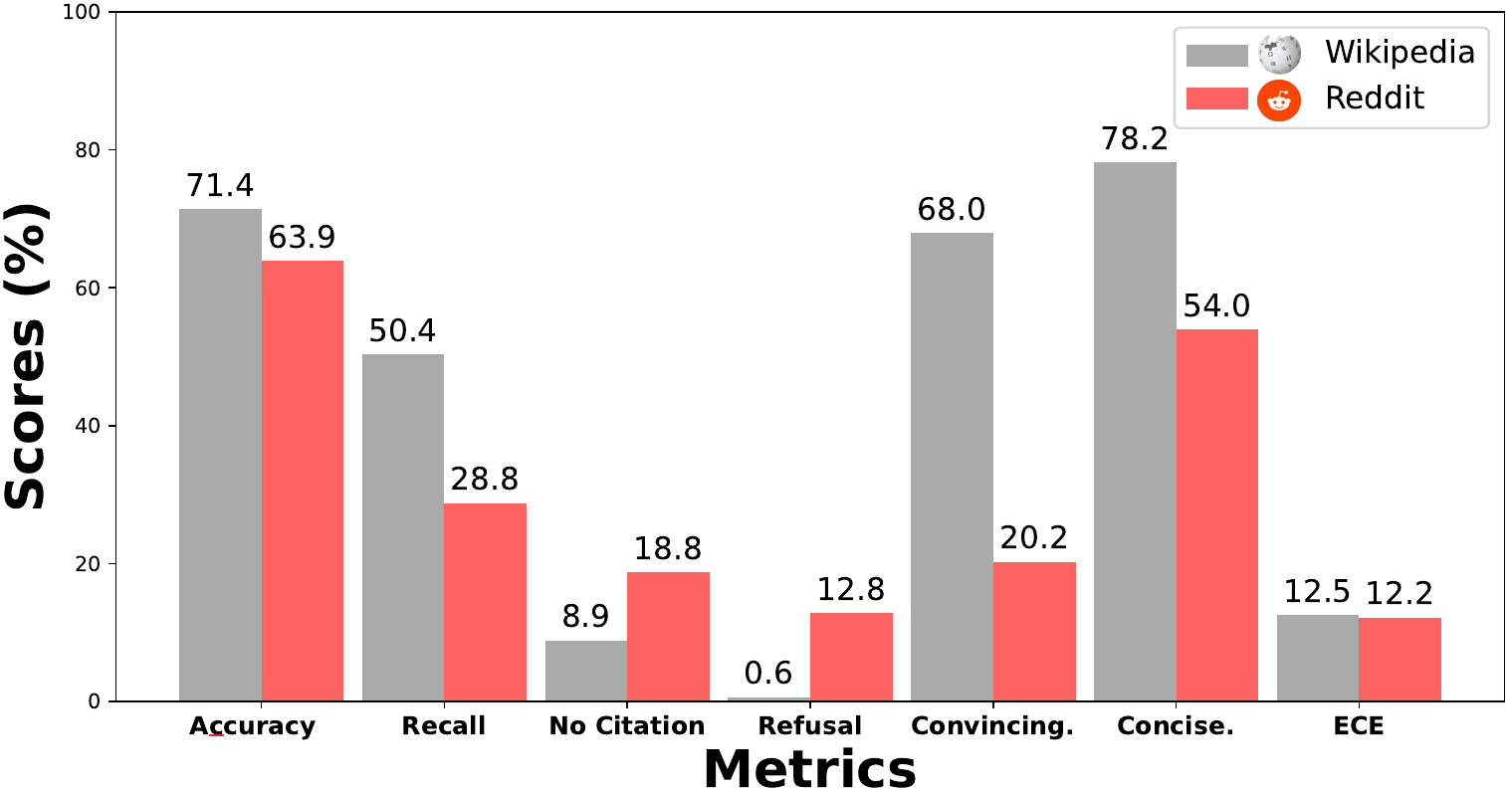}
  \caption{Results on Wikipedia and Reddit dataset. We rescaled each metric to a 0\%-100\% range. }
  \label{fig:rdt}
\end{figure}

We use GPT-4o to imitate the style of Reddit posts and substitute the documents to generate a set with less convincing documents. We fix other settings and use this dataset to evaluate the model's behavior under the situation with a less convincing knowledge source. We show the results of our experiment on Llama-3.1-8B-Instruct in Figure \ref{fig:rdt}.

Our results show that when external documents are less convincing, the model suffers a significant drop in performance, especially on recall, and is more likely to generate a refusal answer even if a ground truth document is provided. This indicates that the quality of external knowledge sources significantly impacts citation generation tasks.

\subsubsection{Tug-of-war between knowledge}
\label{tug}
The tug-of-war between external and prior knowledge in RAG, especially in conflict scenarios, has been widely studied \citep{wu2024clasheval, jin2024tugofwarknowledgeexploringresolving}. Our results show that behavior happens in citation generation. We dive into two different types of scenarios where the difficulties of the question or the level of knowledge grasp differ. The distribution of the dataset is shown in Figure \ref{fig:pie}.

\begin{figure}[H]
  \includegraphics[width=\columnwidth]{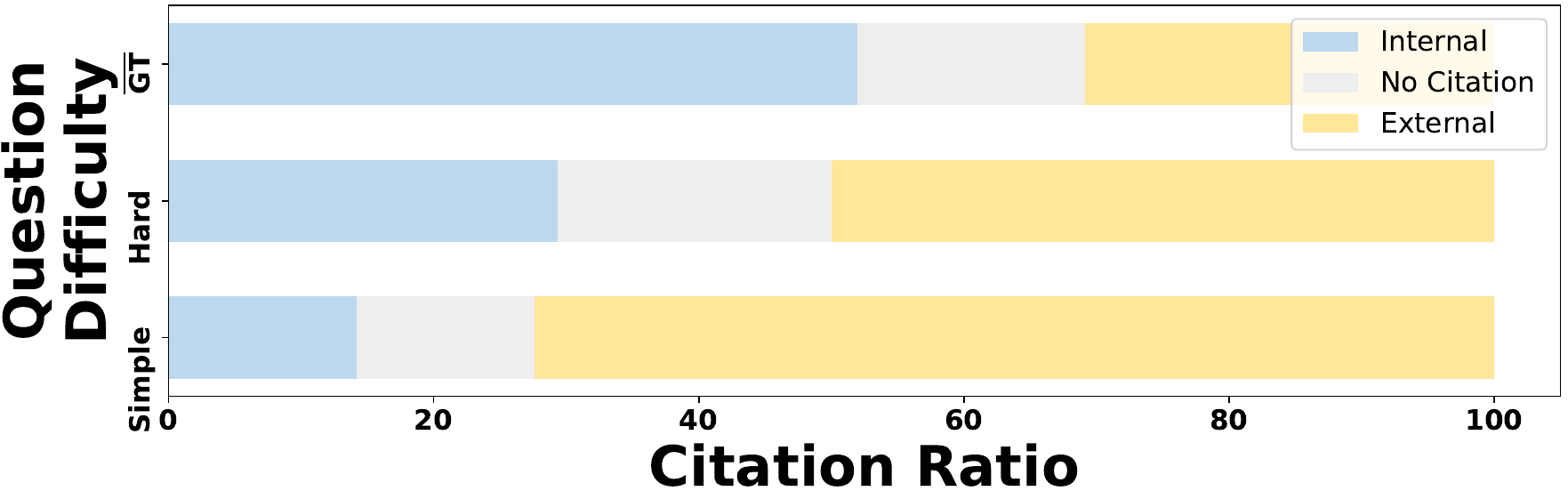}
  \caption{Citations questions with different difficulties in leveraging external knowledge.}
  \label{fig:t2}
  \vspace{-5mm}
\end{figure}

As shown in Figure \ref{fig:t2}, we separate the dataset into (1) with documents that exactly contain the answer string (Simple); (2) with documents that entail the answer but do not contain the answer string (Hard); (3) $\overline{\textbf{GT}}$. We find the model prefers citing internal knowledge when external documents become harder to leverage.

\begin{figure}[H]
  \includegraphics[width=\columnwidth]{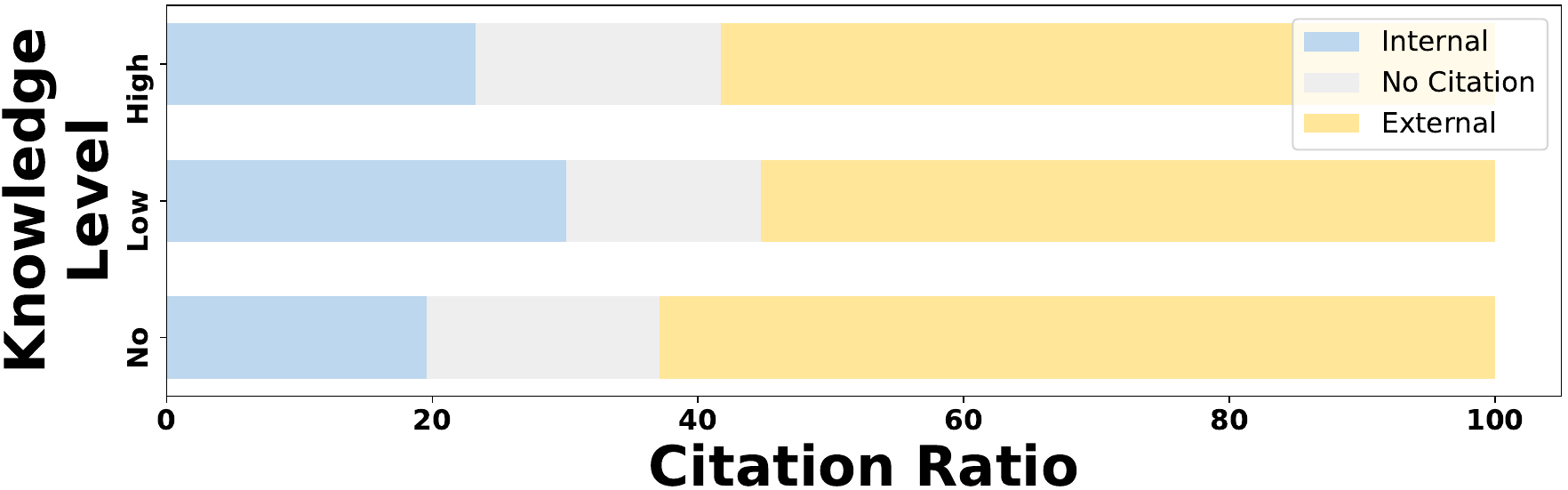}
  \caption{Citations for questions that the Llama3.1-8B has different levels of knowledge.}
  \label{fig:t1}
\end{figure}

We separate the dataset by the model's knowledge level to the question into three: Without Knowledge, Low and High-level grasp. The model prefers to cite external knowledge when it has no knowledge about the question and when its knowledge level is high, as demonstrated in Figure \ref{fig:t1}.


The tug-of-war between knowledge suggests that the model should rely more on external references when the question is straightforward, and the model's knowledge is limited while leveraging prior knowledge more if the question is complex.

\subsubsection{Dishonest internal reference generation}
When external documents are not convincing, the model may rewrite them and claim them as internal knowledge. We believe this reflects dishonesty when generating internal references, but we do not penalize this behavior since high-quality rewriting would be more helpful to users. To study this behavior, we define \textbf{plagiarism}, which refers to the case as internal reference $\textcolor{inblue}{R_i}$ entails the answer $a$ for questions in the \textbf{GT}, $\overline{\textbf{PK}}$ set. The plagiarism rate is $
\text{PR} = \frac{1}{N}\sum_{i = 0}^ N \frac{1}{m_i}\sum_{j = 1}^{m_i} \phi(\textcolor{inblue}{R_{ij}^{in}}, a_i)$, and the severity is $\text{PS} = \frac{1}{N}\sum_{i = 0}^ N \frac{1}{m_i}\sum_{j = 1}^{m_i} \textcolor{confgreen}{P_{ij}}\cdot \phi(\textcolor{inblue}{R_{ij}^{in}}, a_i)$.

$N$ is the size of the \textbf{GT}, $\overline{\textbf{PK}}$ set and $m_i$ is the number of internal references for the $i$-th sample. $\textcolor{inblue}{R_{ij}^{in}}$ and $\textcolor{confgreen}{P_{ij}}$ are the $j$-th internal reference of the $i$-th sample in the \textbf{GT}, $\overline{\textbf{PK}}$ set. PR indicates the proportion of plagiarized references, while PS represents the average confidence falsely reported. We show PR and PS over 3 methods in Table \ref{tab:plagiarism}.

\begin{table}[ht]
\centering
\small

\begin{tabular}{lcccc}
\toprule
& \multicolumn{2}{c}{Llama3.1-8B}     & \multicolumn{2}{c}{Llama3.1-70B}\\
\cmidrule(lr){2-3} \cmidrule(lr){4-5} 
           & PR $\downarrow $  & PS  $\downarrow$   & PR $\downarrow$    & PS$\downarrow$   \\ 
\midrule
\textsc{PostCite} & 0.154 & 0.124 & 0.128 &0.101\\
\textsc{Guided-RAE}   & 0.143 & 0.119 & 0.111 &0.92\\
Ours       & \textbf{0.054} & \textbf{0.049} & \textbf{0.105} &\textbf{0.076}\\
\bottomrule

\end{tabular}

\caption{PR and PS for different methods.}
\label{tab:plagiarism}
\end{table}

We identify the presence of plagiarism, and our methods demonstrate a relatively lower plagiarism rate and plagiarism severity, indicating the effectiveness of the alignment process.

Besides, we find that the larger model is worse than the smaller model with \fname{}. The phenomenon that a larger model has a higher plagiarism rate might indicate that the model learns more about the preference for convincingness and conciseness and becomes more inclined to rewrite the external documents when the quality of the external documents is not satisfying.

\subsection{Ablation of our method}
\label{ablation}

In addition to the Guided-\fname{}, we provide results of extra ablations, including (1) Directly generating references and answers without using our \fname{} paradigm; (2) Our alignment algorithm without \fname{} paradigm; (3) Our alignment algorithm without weighted loss. The results in Tables \ref{ablation 8B} and \ref{ablation 70B} are presented, along with an analysis below.

\definecolor{mygray}{gray}{0.8} 
\renewcommand{\arraystretch}{0.8} 
\begin{table}[ht]
\centering
\small
    \resizebox{\columnwidth}{!}{%
\begin{tabular}{p{1.5cm}lccccc}

\toprule
Scenario & Method        & Acc   & Recall     & Conv. & Conc. & ECE \\
\bottomrule
\addlinespace[0.2em]
\multirow{5}{*}{\fbox{\textbf{GT}, \textbf{PK}}}&Direct & 68.03 & 30.45   &  3.44  & 3.70  & 0.17  \\
&Ours w/o \fname{}    & 75.87 & 31.56      &  3.54  & 3.76  & 0.13  \\

&Ours w/o Weight & 71.43 & 50.36      & 3.61 & 3.99  & 0.14  \\

&Ours          & \textbf{75.90} & \textbf{63.72}      & \textbf{3.61} & \textbf{4.05}  & \textbf{0.10}  \\
\bottomrule
 
\addlinespace[0.2em]
\multirow{5}{*}{\fbox{\textbf{GT}, $\overline{\textbf{PK}}$}}&Direct & 53.90 & 25.11      &   3.50   &   3.65    &    0.18   \\
&Ours w/o \fname{}    & 64.29 & 22.03      &   3.52   &    3.72   &    0.11   \\

&Ours w/o Weight  & 69.05 & 40.17      & 3.62 & \textbf{3.85}  &   0.14  \\

&Ours          & \textbf{69.05} & \textbf{51.84}      & \textbf{3.64} & 3.78  & \textbf{0.11}  \\
\bottomrule
\addlinespace[0.3em]
\multirow{5}{*}{\fbox{$\overline{\textbf{GT}}$, \textbf{PK}}}&Direct & 36.40  & 29.17      &  3.41    &   3.20    &   0.20    \\
&Ours w/o \fname{}    & 61.38 & 52.79      &   3.65   &    3.67  &    0.13   \\

&Ours w/o Weight & 62.24 & 49.80      & \textbf{3.66} & 3.55  &     0.12  \\

&Ours          & \textbf{62.64} & \textbf{59.80}      & 3.65 & \textbf{3.83}  & \textbf{0.11}  \\
\bottomrule
\end{tabular}}
\caption{Ablation results on Llama3.1-8B-Instruct}
\label{ablation 8B}
\end{table}

\begin{table}[ht]
\centering
\small
    \resizebox{\columnwidth}{!}{%
\begin{tabular}{p{1.5cm}lccccc}

\toprule
Scenario & Method        & Acc   & Recall     & Conv. & Conc. & ECE \\
\bottomrule

\addlinespace[0.2em]
\multirow{5}{*}{\fbox{\textbf{GT}, \textbf{PK}}}&Direct   &  78.82   &  40.24  &  3.48   &   4.21  &  0.22 \\
&Ours w/o \fname{}  &  72.01  &   58.32  &  3.50    &   3.92    &   0.12    \\

&Ours w/o Weight  & 85.71 &  58.30   &  3.65    &   4.02    &   0.10    \\

&Ours           & \textbf{85.72} &  \textbf{78.79}   &    \textbf{3.69}  &  \textbf{4.42}     &   \textbf{0.10}     \\
\bottomrule  

\addlinespace[0.2em]
\multirow{5}{*}{\fbox{\textbf{GT}, $\overline{\textbf{PK}}$}}&Direct    &  61.71   &    42.76  &  3.50   &   4.19  &  0.22\\
&Ours w/o \fname{}    & 65.97 &  53.77   &   3.54   &    4.07   &    0.15   \\

&Ours w/o Weight  & \textbf{71.88} &  53.38   &  3.44    &   3.98   &    \textbf{0.12}    \\

&Ours          & 71.25 &  \textbf{60.32}   &  \textbf{ 3.60}   &     \textbf{4.47}  &    0.13   \\
\bottomrule
\addlinespace[0.2em]
\multirow{5}{*}{\fbox{$\overline{\textbf{GT}}$, \textbf{PK}}}&Direct  & 47.80 &  42.99   &   3.49   &    4.10   &  0.15  \\
&Ours w/o \fname{}    & 62.48 &  60.12   &   3.46   &    4.25   &    0.12   \\

&Ours w/o Weight  & 75.11 &  72.45   &  \textbf{3.67}    &   4.10    &   0.12    \\

&Ours           &  \textbf{75.64}&  \textbf{75.71}  &    3.67  &   \textbf{4.42}   &   \textbf{0.09}     \\
\bottomrule
\end{tabular}}
\caption{Ablation results on Llama3.1-70B-Instruct}
\label{ablation 70B}
\end{table}

RAEL and weighted loss significantly improve the citation recall, convincingness and conciseness. The accuracy is also slightly improved. The result aligns with our focus on citations in the training and data creation process.

The training without weights sometimes achieves comparable or better results than the weighted training, but that does not happen very often. Considering the low extra cost and convenience of integrating weighted loss, it is still a beneficial way to improve overall performance. We also notice this case happens often in the $\overline{\textbf{PK}}$ set (with questions for which LLMs have no internal knowledge), which means the weighted loss might be more useful for improving internal citation quality.

\vspace{-8mm}
\vspace{5mm}
\section{Human Evaluations}

\label{human}

We design a webpage for human evaluations, as shown in Figure \ref{fig:human page}, and conduct human evaluations to justify our automatic metrics. Three participants with fluent English proficiency took part in the evaluation, each of whom underwent preliminary testing and was assigned 100 examples. 

\textbf{NLI Accuracy.} Evaluations demonstrate that our accuracy metric has only a 2.65\% False Positive Rate and 6.19\% False Negative Rate.

\textbf{Convincingness aligns with trustworthiness, and Conciseness aligns with verification difficulty.} The participants were required to rate the level of trustworthiness and difficulties in verification for certain generated references. We use Pearson Correlation Coefficients (PPCs) as an indicator of the correlation. Table \ref{tab:humantable} implies that the correlation between human judgment and automatic evaluations is similar to that between individuals.

\vspace{-2mm}
\renewcommand{\arraystretch}{0.8}
\begin{table}[H]
    \centering
    \small
    \begin{tabular}{ccccc}
    \midrule
          Evaluator &    Convincingness& Conciseness & \\
           \midrule
         Automatic  &  0.53  & 0.66  & \\
         Individual $\beta$  & 0.58 &  0.74  & \\
         \bottomrule
    \end{tabular}
    \caption{PPCs between individual $\alpha$ and two evaluators: (1) Our automatic evaluator and (2) Individual $\beta$.}
    \label{tab:humantable}
\end{table}

\vspace{-3mm}
\section{Conclusion}

In this paper, we present a \textcolor{exgold}{Context}-\textcolor{inblue}{Prior} Augmented Citation Generation task, requiring LLMs to generate citations and fine-grained references from both external contexts and internal parameter knowledge. Our comprehensive evaluation of answer helpfulness, citation faithfulness, and reference trustworthiness reveals the challenge for LLMs to generate trustworthy citations.

We also propose \fname{} paradigm for this task and a method, \mname{}, to unleash the model's capacity to cite parameter knowledge with trustworthiness, facilitating a more transparent citation generation. In-depth studies reveal the significance of the quality of knowledge sources and highlight the LLM's selective utilization of external and internal knowledge in citation generation.

\section{Limitations}
Using more open-source models can still enrich our experiments, as different parameter knowledge in the LLMs makes a huge difference in the data sampling and dataset-splitting process. 

While we believe our datasets closely reflect the distribution found in real-world scenarios, possible bias may still be introduced during reranking and selection in the data sampling process. We also fabricate Reddit-style data from Wikipedia, which may not represent an authentic low-quality knowledge source. We still observe minor correct answers less than 5\% in the $\overline{\textbf{GT}}$, $\overline{\textbf{PK}}$ set, implying that our annotation process still has some omissions. 

Although our method is effective, we still leave room to consider and explore the internal mechanism of parameter knowledge utilization, and future works may focus on the intended control of citing external or internal knowledge.

\section{Ethical Considerations}
We require large models to memorize parameter knowledge, which could raise copyright issues, as some of the data used for training may be copyrighted. However, the complete memorized knowledge does not exceed 200 words per article. Our experimental results do not need full access to the memorized content generated in the model's intermediate steps, and we only analyze the final cited span, which is less than 50 words.

Our dataset includes Reddit-style rewritings of Wikipedia content, which might contain inaccurate or misleading information. We ensure this dataset is private only for our supplementary experiment and not publicly available.

Three non-paid volunteers participated in our human evaluation. We ensured that all participants were informed about the research objectives, the tasks involved, and their role in the evaluation process. Their involvement was entirely voluntary. Volunteers gave their consent and had the right to withdraw at any time without any consequences. We ensured the anonymity and confidentiality of all personal data in the evaluation.

\bibliography{custom}

\newpage
\appendix

\section{Related Work}
\label{rw}
\textbf{LLM Citation Generation.}
ALCE \citep{gao-etal-2023-enabling} introduced a paradigm for citation generation in LLMs and established key evaluation metrics. Subsequent work improved citation quality by refining granularity \citep{xu2024aliiceevaluatingpositionalfinegrained, zhang-etal-2024-towards-fine-grained}, enhancing model attribution evaluation \citep{yu-etal-2024-revealing}, and exploring user-centered effectiveness measure \citep{worledge2024extractiveabstractivespectrumuncoveringverifiability}. Recent advances follow two approaches: fine-tuning models \citep{huang-etal-2024-learning, ye-etal-2024-effective,huang-etal-2024-training, li-etal-2024-improving-attributed} and designing structured pipeline \citep{DBLP:journals/corr/abs-2404-03862, lee2023towards, xu2024searchinthechain, gao-etal-2023-rarr, ding2024attentiondependencyparsingaugmentation, fierro-etal-2024-learning, sun-etal-2024-towards-verifiable}. Despite these efforts, existing studies have not considered integrating LLMs' internal and external knowledge with confidence quantification.

\textbf{LLM Parameter Knowledge.}
Some studies leverage parameter knowledge to enhance generation and attribution \citep{yu2023generate, sun2023recitationaugmented}, while others explore the model's behavior in different scenarios where parameter knowledge and contextual knowledge conflict \citep{minder2024controllablecontextsensitivityknob, ming2024faithevallanguagemodelstay, cheng2024understandinginterplayparametriccontextual}. However, these studies have not fully addressed the adaptive use of parameter and contextual knowledge in large models, nor have they provided sufficient interpretability in utilizing parameter knowledge.

\textbf{Trustworthiness and factuality of LLMs.}
Some studies have focused on human-centered LLMs, considering the trustworthiness and verifiability of model-generated content \citep{venkit2024searchenginesaiera, ding2025citationstrustllmgenerated}. However, these approaches lack quantitative definitions and targeted improvements. To improve the factuality issues in LLM outputs, some studies have developed fact verification methods to detect hallucinations \citep{min2023factscorefinegrainedatomicevaluation, niu-etal-2024-veract, scirè2024truthmirageendtoendfactuality}, which have also been used to calibrate model output confidence \citep{fadeeva2024factcheckingoutputlargelanguage, yuan2024factlevelconfidencecalibrationselfcorrection}. Meanwhile, studies have also explored the model's confidence of the generated content in RAG or plain QA tasks \citep{ozaki2024understandingimpactconfidenceretrieval, tao-etal-2024-trust}. However, these works have not considered integrating citation tasks to enhance interpretability.


\section{Human Evaluation}
\label{sec:appendix}
\begin{figure}[H]
    \centering
    \includegraphics[width=\columnwidth]{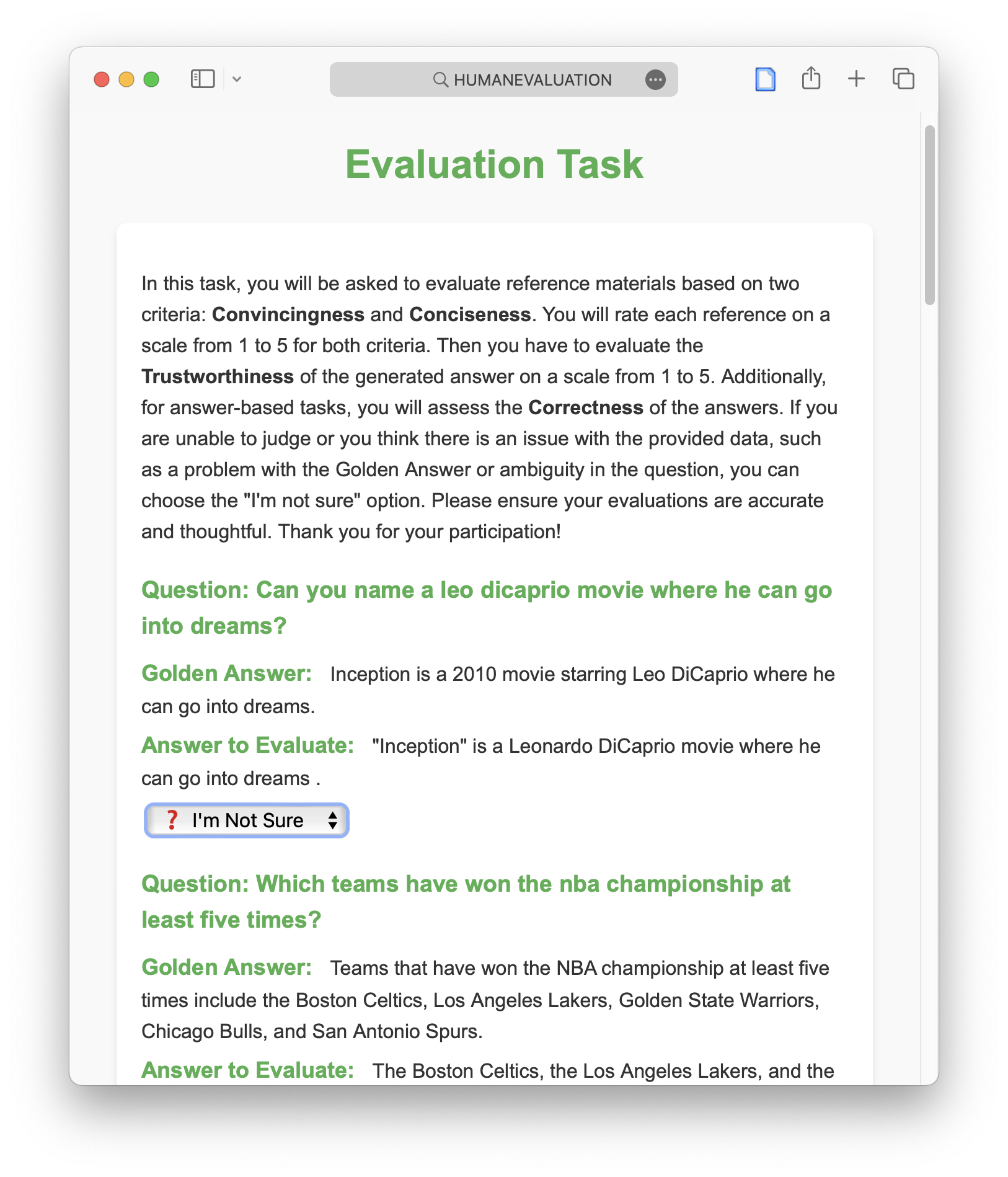}
    \caption{Webpage for human evaluations}
    \label{fig:human page}
\end{figure}

We designed a webpage for conducting human evaluations, as shown in Figure \ref{fig:human page}. For reference evaluation tasks, each participant is asked to evaluate the Convinvingness and conciseness of a given reference. For answer evaluation tasks, each is asked to evaluate the correctness of the answer.

\textbf{NLI Accuracy is better than Exact Match.} 
Previous citation generation tasks use String Exact Match to compute accuracy. However, it is difficult to exhaust all possible answers, and the response may still mention the golden answer even if the intended answer is the other. To alleviate the problems, we desgin our NLI-based Accuracy.

We asked our participants to annotate the answer manually and calculate the False Positive Rate and False Negative Rate of the results from NLI Accuracy and Exact Match. We show the FP Rate, FN Rate and Accuracy in Table \ref{acceval}, which demonstrates our method has relatively fewer mistakes and a higher overall accuracy.

\begin{table}[ht]
\centering
\renewcommand{\arraystretch}{1.5} 
\small

\begin{tabular}{cccc}
\hline
             & FP Rate $\downarrow$ & FN Rate $\downarrow$ & Accuracy $\uparrow$ \\
\hline
String Exact Match & 3.54\%               & 11.24\%              & 83.11\%            \\
Ours        & \textbf{2.65\%}               & \textbf{6.19\%}              & \textbf{91.27\%}            \\
\hline
\end{tabular}
\caption{Comparison of Exact Match and our method for accuracy evaluation}
\label{acceval}
\end{table}

\begin{figure}[ht] \centering \includegraphics[width=\columnwidth]{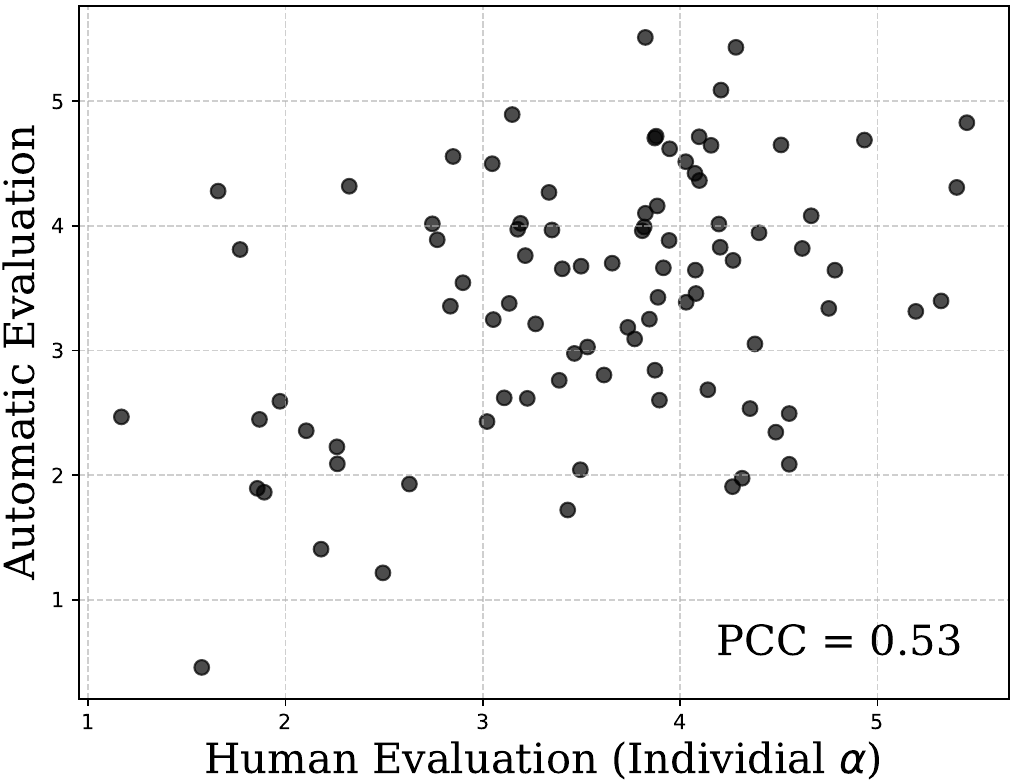} \caption{Scatter plot for Convincingness Evaluation. Gaussian noise $\mathcal{N}(\mu = 0, \sigma^2 = 0.5)$ has been added to the points to prevent overlap.} \label{fig:conv} \end{figure}

\begin{figure}[ht] \centering \includegraphics[width=\columnwidth]{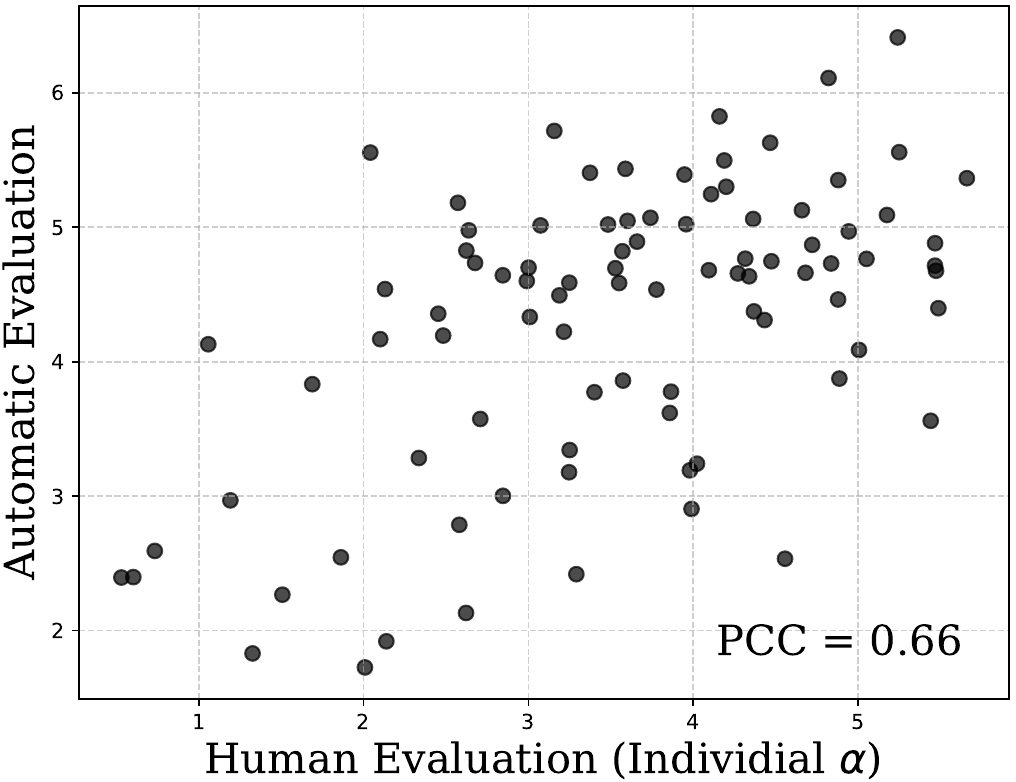} \caption{Scatter plot for Conciseness Evaluation. Gaussian noise $\mathcal{N}(\mu = 0, \sigma^2 = 0.5)$ has been added to the points to prevent overlap.} \label{fig:conc} \end{figure}

\textbf{Convincingness and Conciseness.}
We visualize the correlation between human evaluations from individual $\alpha$ and our GPT-4o-mini automatic evaluator in Figures \ref{fig:conv} and \ref{fig:conc}, which shows the correlation on Convincingness and Conciseness, respectively.

In both plots, the PCC values highlight the correlation between human ratings and the automatic evaluation method.

\section{Robustness to Shortcut Cases}
\label{app:robust}
We discuss three possible shortcut cases: (1) Cite all the documents provided to ensure the Recall score. In this case, references will suffer from high redundancy. (2) Only cite the minimum span to ensure conciseness; the convincingness of the reference will be relatively lower. (3) Only cite external documents or parameter knowledge. In this case, the accuracy metric will significantly decrease on a dataset containing questions the LLM has no knowledge about or documents containing the answer is not provided.

\section{Datasets}
\label{app:ds}
\begin{figure}[t]
  \includegraphics[width=\columnwidth]{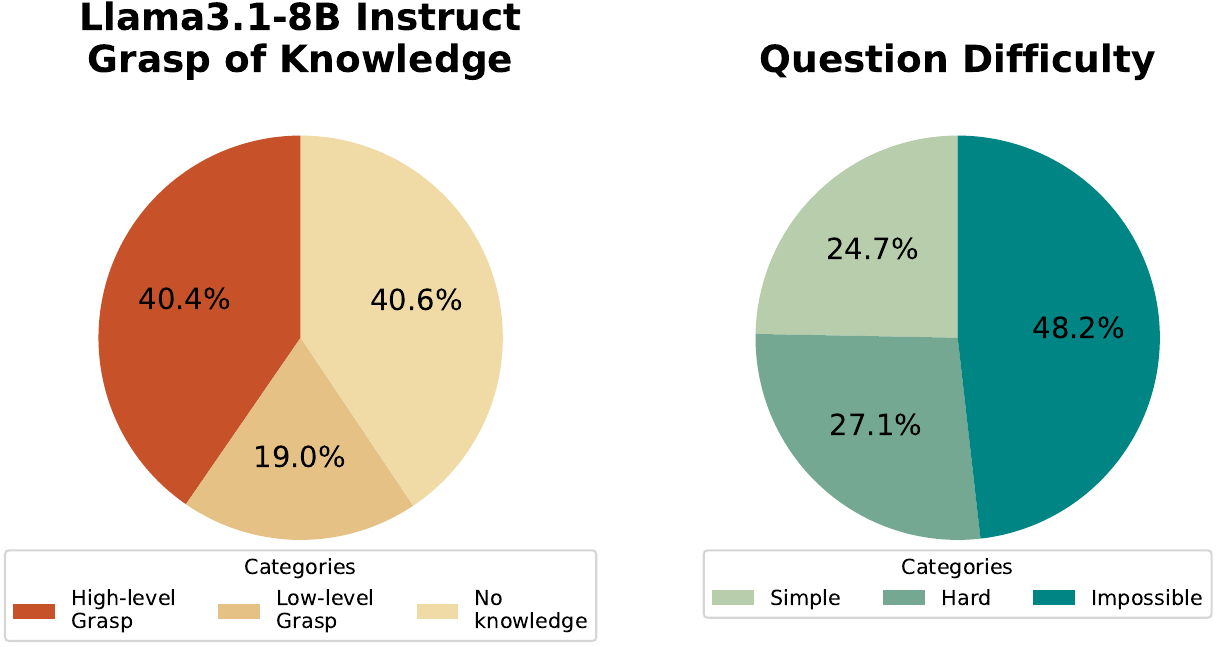}
  \caption{Dataset Distribution}
  \label{fig:pie}
\end{figure}

The datasets we used focus on various types of QA, multi-document reasoning, and internal/external knowledge fidelity, making them suitable for comprehensively evaluating the model's overall performance on our task. They are all factual questions from the real world, and relevant knowledge can be retrieved from Wikipedia. The distribution of the dataset according to the partitioning method in \ref{tug} is in Figure \ref{fig:pie}, and examples are in Table \ref{tab:dataset examples}.

\begin{table*}[h!]
    \centering
    \small
    \begin{tabular}{clp{9cm}p{1.2cm}}
    \toprule
    \textbf{Dataset} & \textbf{Category} & \textbf{Example} & Amount\\
    \midrule
      \multirow{14}{*}{CRAG}
       & Simple &Which movie won the Oscar Best Visual Effects in 2021?&580 \\
         \cmidrule(lr){2-4} 
       & Simple with condition & What is a movie to feature a person who can create and control a device that can manipulate the laws of physics?&260\\
                \cmidrule(lr){2-4} 
         & Multi-hop & How long is the longest river in Alabama?& 191\\
                  \cmidrule(lr){2-4} 
         & Comparison &Which movie was created first, A Walk to Remember or The Notebook? &123\\
                  \cmidrule(lr){2-4} 
         & False premise &When did Hamburg become the biggest city in Germany?& 100 \\
          \cmidrule(lr){2-4} 
         & Post-processing & How many 3-point attempts did Steve Nash average per game in seasons he made the 50-40-90 club?& 48\\
          \cmidrule(lr){2-4} 
         & Aggregation & How many family movies were there that came out in 1994? & 220\\

         \midrule

        \multirow{19}{*}{FRAMES} & Temporal reasoning & Was the person who served as president of the Scottish National Party from 1987 to 2005 alive when the party was founded?& 49\\
                  \cmidrule(lr){2-4} 
         & Multiple constraints & As of August 3, 2024, which rabbi worked for both Reform Congregation Keneseth Israel in Philadelphia and Congregation Beth Israel in West Hartford, Connecticut?& 124\\
                   \cmidrule(lr){2-4} 
         & Numerical reasoning & What painting was stolen from The Louvre exactly 56 years before the birth of activist and songwriter Serj Tankian?& 56\\
                   \cmidrule(lr){2-4} 
         & Tabular reasoning & What is the birthplace and hometown of the winning goal scorer of the 2010 Vancouver Olympics, Men's Ice Hockey event? &42\\
                   \cmidrule(lr){2-4} 
         &  Post processing& This athlete was the first man to run the 100 meters in under 10 seconds at an Olympic Games. Which NFL team was he drafted by? &20\\
         \midrule

        \multirow{12}{*}{SFE} &Years & In which year did Godwin Obaseki switch from the APC to the PDP? & 480\\
                           \cmidrule(lr){2-4} 
        & Names & Who is the brother of southern gospel music singer Lynda Tait Randle that is associated with the musical groups DC Talk and Newsboys& 104\\
                           \cmidrule(lr){2-4} 
        & Locations & Which city is the Saint Nicholas Monastery, an Eastern Orthodox monastery that was made the seat of the Eastern Orthodox Eparchy of Mukachevo in 1491, located in? & 142\\
                           \cmidrule(lr){2-4} 
        & News & How much is Sheldon Rankins' contract with the Cincinnati Bengals worth in millions of US dollars as agreed upon on March 30, 2023? & 299\\
                 \bottomrule
    \end{tabular}
    \caption{Examples for each category in the dataset}
    \label{tab:dataset examples}
\end{table*}

\textbf{CRAG} \citep{yang2024crag} is a factual question-answering benchmark of 4K question-answer pairs. CRAG is designed to encapsulate diverse questions across 5 domains and 8 question categories, as indicated in Appendix \ref{app:ds}. To prevent conflicts between external documents and the LLM's internal knowledge due to knowledge updates, we have removed time-sensitive data, retaining only questions categorized as \textit{static}.

\textbf{FRAMES} \citep{krishna2024factfetchreasonunified} is a high-quality evaluation dataset designed to test LLMs’ ability to provide factual responses and evaluate the reasoning required to generate final answers. Each question in FRAMES requires reasoning with 2-15 Wikipedia articles. We select all the questions that require exactly 2 Wikipedia articles to facilitate the document annotation step.

\textbf{SituatedFaithfulnessEval} \citep{huang2024enhancinglargelanguagemodels} benchmarks LLM's ability to demonstrate situated faithfulness, dynamically calibrating their trust in external information based on their confidence in the internal knowledge and the external context. We select the factual QA questions from the \textbf{ClashEval} \citep{wu2024clasheval} subset, which is used to quantify the tug-of-war between an LLM’s internal prior and external evidence.

For annotation, we first combine the three datasets, and for each question in the combined dataset, our pipeline first retrieves top-100 passages from a chunked Wikipedia snapshot using a GTR retriever \citep{wang2021retrieving} and the question as the query. Then, we apply an NLI model \citep{honovich-etal-2022-true-evaluating} is applied to annotate documents. For each document $d$ and the answer $a$, if $\phi(d, a) = 1$, then the document is annotated as ground truth. After filtering out the questions without ground truth documents retrieved, we make two data points for each question with different settings, one with a random number of ground truth documents and the other without any ground truth document. Finally, we supplement retrieved documents with $\phi(d, a) \ne 1$ as irrelevant documents so that each data point is paired with exactly 5 documents. According to whether it contains a \textbf{G}round \textbf{T}ruth document, we split each datapoint into 2 settings \textbf{GT} (with ground truth documents) and $\overline{\textbf{GT}}$ (without ground truth documents).

\section{Implementation Details}

\textbf{NLI model.} We apply \href{https://huggingface.co/google/t5_xxl_true_nli_mixture}{TRUE} as the NLI model, which returns a bool value $\phi(\text{premise}, \text{hypothesis}) = 1$ if the premise entails the hypothesis.

\textbf{Training setup and Hyper Parameters}
We use LoRA \cite{hu2022lora} to fine-tune our models with $\text{rank} = 8$ and learning rate as $\text{lr} = 10^{-4}$. We train 2 epochs on our dataset in total. For GPT-4o and GPT-4o-mini used as baselines and evaluation models, we use $\text{temperature} = 0.5$ and $\text{top\_p} = 0.9$.

\subsection{Baselines}
\label{app:baseline}
We detail the baselines that we used in our experiments below.

\textbf{Guided-\fname{}}
We use a two-shot prompt to guide the model in using our Rational Attribution and Elaboration framework.

\textbf{\textsc{FootNote.}}
We asked the LLM to generate an answer with reference the \verb| \footnote| format following each sentence. We use \verb|\footnote[confidence]{reference}| to represent \textcolor{inblue}{internal citation} and use \verb|\footnote[idx]{reference}| to represent \textcolor{exgold}{external citation}, where \verb|idx| is the external document index.

\textbf{\textsc{PostCite.}}
We prompt the LLM to generate a response according to retrieved documents and segment the response into sentences. For each sentence, we cite the retrieved document with the highest score using a GTR retriever if the score exceeds a given threshold $\eta$; otherwise, we ask LLM to generate a document with a confidence score to cite as an internal citation. In our implementation, we dynamically set $\eta$ to ensure the same total \textcolor{exgold}{external} and \textcolor{inblue}{internal citations}.

\textbf{\textsc{Recitation Augmented Generation.}}
\citeposs{sun2023recitationaugmented} work utilizes parameter knowledge to augment LLM Generation. This baseline sample $k$ passages from LLM's parameter knowledge and generates $k$ corresponding answers given each passage as context. The final answer is determined through majority voting. We adopt a logits-based method, CCP \citep{fadeeva2024factcheckingoutputlargelanguage}, to obtain a confidence score for each generated passage. In our implementation, we set $k$ to 5. The answer is not a single word, so we are unable to use string match to realize majority voting, so we assume answer $a_1$ is the same as $a_2$ if $\phi((q;a_1), (q;a_2))$ or $\phi((q;a_2), (q;a_1))$, and classify all the answers. The final answer is randomly chosen from the largest set.

\textbf{\textsc{Front.}}
\textsc{Front} is a training-based baseline designed to enhance citation quality through fine-grained grounded citations \citep{huang-etal-2024-learning}. The model first extracts supporting quotes from retrieved documents and uses them to guide the answer-generation process, ensuring precise and grounded responses. Then, it further optimizes the consistency between the grounded quotes and the generated answers using preference optimization techniques. In our implementation, we use our dataset to generate data and train the model following the pipeline of \textsc{Front}, replacing our \fname{} paradigm and \mname{}.

\textbf{\textsc{ContextCite.}}
\textsc{ContextCite} is a context attribution baseline that identifies the specific parts of the context responsible for generating a model's response \citep{cohen-wang2024contextcite}. \textsc{ContextCite} emphasizes contributive attribution, and it achieves this through a surrogate model trained to approximate how excluding or including specific context sources affects the response. \textsc{ContextCite} utilizes sparse linear modeling to provide efficient and scalable attribution, ensuring that each identified source significantly impacts the generated output. We regard the attributed part in the context as the reference.

\section{Abstention}
\label{abstention}
When LLM has no knowledge about a question and no ground truth documents are provided, the golden answer is a refusal answer. When the answer contains any of the pre-listed refusal sentences or when the NLI model computes \(\phi(a, \text{"Unable to answer."})=1\), we consider the model to have given a refusal answer. We measure the rate of refusal in the \colorbox{gray!10}{$\overline{\textbf{GT}}$, $\overline{\textbf{PK}}$} set of different baselines and our \mname{}, as shown in Table \ref{maintable2}

\textbf{Our method allows the model to abstain when necessary.} When no ground truth documents are provided and the model also has no knowledge about the question, the model trained on our methods will be more likely to abstain from answering. This behavior allows the model to give a more trustworthy response and reduce hallucinations.

\begin{table}[h!]
    \centering
    \small 
    \resizebox{\columnwidth}{!}{%
    \begin{tabular}{cccc}
        \toprule

 &\multicolumn{2}{c}{Llama3.1-Instruct}    \\
 \cmidrule(lr){2-3}
 &8B&  70B&  GPT-4o \\ 
 \midrule
 
 \textsc{Recite$^\dagger$}&1.53 & 1.97& - \\ 
       
\textsc{Front$^\ddagger$}&2.55 & 3.67& -  \\ 

\textsc{ContextCite$^\ddagger$}& 24.7 & 22.0 &-\\ 

\textsc{FootNote}& 18.4 &16.19  &10.32\\ 

\textsc{PostCite}& 21.4 & 3.71 & 17.42\\ 

\textsc{Guided-RAE}&6.8 &6.1 & 18.7  \\ 

Ours& \textbf{28.84}& \textbf{28.21}& - \\ 

        \bottomrule
    \end{tabular}}
    \setlength{\fboxsep}{1pt}
    \caption{Refusal Rates on test set \colorbox{gray!10}{$\overline{\textbf{GT}}$, $\overline{\textbf{PK}}$}.}
    \label{maintable2}
\end{table}

\section{Prompts}
\label{prompts}
We show our prompts for evaluating Convincingness, Conciseness and the prompt for generating \fname{} in Multi-scenario Trustworthy Data sampling in Figure \ref{fig:conv prompt}, \ref{fig:conc prompt}, and \ref{fig:gen prompt}, respectively.

\begin{figure*}[!thp]
    \centering
    \begin{mdframed}[innerrightmargin=20pt]
  \begin{CustomVerbatim}
You are an evaluator tasked with assessing the Convincingness of a text. Convincingness is defined as the text's ability to avoid raising doubts about its truthfulness in the reader. Consider the following criteria while scoring: 

1. **Logical Consistency**: Evaluate whether the text avoids logical errors or contradictions.
2. **Subjectivity**: Assess whether the language is objective and free from excessive bias or personal opinions.
3. **Coherence and Focus**: Determine if the arguments are well-connected and focused rather than scattered or overly parallel.
4. **Information Density**: Consider whether the text provides sufficient relevant information to substantiate its claims.

Please assign a score between 1 and 5 based on the following detailed guidelines:

1. **Score: 1 (Very Low Convincingness)**  
   - Contains multiple logical errors or glaring contradictions.  
   - Dominated by subjective or emotional language.  
   - Arguments are highly scattered, with no clear connections between points.  
   - Lacks sufficient information to support its claims.

2. **Score: 2 (Low Convincingness)**  
   - Contains some logical inconsistencies or weak reasoning.  
   - Has a noticeable bias or subjective tone.  
   - Arguments are somewhat scattered, with limited connections between points.  
   - Provides insufficient evidence or relies on vague statements.

3. **Score: 3 (Moderate Convincingness)**  
   - Mostly logical with minor inconsistencies.  
   - Language is somewhat balanced but may lean towards subjectivity.  
   - Arguments are somewhat connected but may lack focus or clarity.  
   - Contains adequate but not robust information density.

4. **Score: 4 (High Convincingness)**  
   - Logically consistent with no major errors.  
   - Language is objective and neutral.  
   - Arguments are mostly coherent and focused.  
   - Provides substantial and relevant evidence for its claims.

5. **Score: 5 (Very High Convincingness)**  
   - Completely free from logical errors or contradictions.  
   - Language is fully objective and professional.  
   - Arguments are tightly connected and maintain a clear focus.  
   - Provides rich, detailed, and highly relevant information to support its claims.

Provide a short explanation of your reasoning, and then output a score between 1 and 5, with formatting like "Score: 3".
   \end{CustomVerbatim}
       \end{mdframed}
    \caption{Prompt for Convincingness Evaluation}
    \label{fig:conv prompt}

\end{figure*}

\begin{figure*}[!thp]
    \centering
    \begin{mdframed}[innerrightmargin=20pt]
  \begin{CustomVerbatim}
You are tasked with assessing the Conciseness of a document sentence by sentence in response to a given question and answer. For each sentence:  
1. Judge whether it positively contributes to answering the question (positive), partially contributes but feels unnecessary or tangential (neutral), or detracts from the relevance (negative).  
2. Provide a brief explanation for your judgment.  

After reviewing all sentences, summarize the overall Conciseness of the document and assign a score between 1 and 5, following these guidelines:  
- **5 (Very High Conciseness):** All sentences are relevant or contribute directly to answering the question.  
- **4 (High Conciseness):** Most sentences are relevant, with a few mildly tangential or unnecessary.  
- **3 (Moderate Conciseness):** A balance of relevant and irrelevant content; reader effort is moderate.  
- **2 (Low Conciseness):** Many sentences are tangential or unnecessary, requiring significant effort to find relevant information.  
- **1 (Very Low Conciseness):** The majority of the document is irrelevant or distracting, with little useful content.  

**Example:**  

**Question:** How many championships has Messi won?  

**Document:**  
1. "Lionel Messi was born on June 24, 1987, in Rosario, Argentina, and is a professional footballer."  
- **Positive:** This sentence establishes Messi as the subject, making it clear the document is on topic.  

2. "His parents are Jorge Messi, a steel factory manager, and Celia Cuccittini, who worked in a magnet manufacturing workshop."  
- **Negative:** This sentence delves into his family background, which feels irrelevant to the question about championships.  

3. "He won his first championship in 2005, leading his team to victory in the U-20 World Cup."  
- **Positive:** This sentence is highly relevant, directly addressing Messi’s championship history.  

4. "His most recent championship was the 2022 FIFA World Cup, where he captained Argentina to victory."  
- **Positive:** This sentence is also highly relevant, discussing a key championship victory.  

5. "Messi hopes to continue playing at a high level and achieve more milestones in his career."  
- **Neutral (slightly negative):** While unrelated to his past championships, it serves as a closing summary and doesn’t significantly detract from the document.  

**Overall Assessment:**  
The document is mostly focused on answering the question, with only one sentence being significantly off-topic. While the fifth sentence is mildly tangential, it serves as a conclusion and does not greatly impact the overall relevance.  

Score: 4 (High Conciseness)  

Provide a short explanation of your reasoning for each sentence, and then output a score between 1 and 5, with formatting like "Score: 3."
   \end{CustomVerbatim}
       \end{mdframed}
    \caption{Prompt for Conciseness Evaluation}
    \label{fig:conc prompt}

\end{figure*}

\begin{figure*}[!thp]
    \centering
    \begin{mdframed}[innerrightmargin=20pt]
  \begin{CustomVerbatim}
You are a Large Language Model with limited knowledge. Given a question, documents, "my knowledge," and a golden answer, please generate a high-quality answer with citation. You should simulate a Large Language Model that thinks step-by-step and outputs references and an answer using the provided documents and "Knowledge in Yourself" (in the "my knowledge section"), but simulate that you cannot see the golden answer. Simulate that you are generating the knowledge yourself, not referring to the "my knowledge" section and the gold answer. 

The response needs to follow the following requirements:
1. Your answer should contain all the information in the golden answer provided (i.e., the golden answer is a subset of your full answer).
2. each statement in your answer should be cited properly, with marks like [1] and [2] to indicate the source of the information. When multiple sources are available, cite a minimum set.
3. Your answer should be concise and contain supporting evidence from the documents provided.

Think step by step to generate the full answer by considering the provided `Documents,` `my knowledge>`, and the golden answer. Here is a guidance:
1. Analyze what kind of knowledge you need to answer the question, and try to find supporting evidence in the documents.
2. Use the provided `Documents` first, and if the information is not enough, use "my knowledge" for a supplement. Scrutinize all the possible "my knowledge" and give an appropriate confidence level according to all the possible "my knowledge."
3. Only use `my knowledge` when provided `Documents` are not sufficient. You don't need to use "my knowledge" for comfirming the information in the provided documents or other unnecessary situations.
4. You pretend to be a Large Language Model with limited knowledge, so you can only use the given documents and "my knowledge" to generate the answer. When using "my knowledge", pretend that you are using the knowledge that you have generated yourself. When thinking about my knowledge, use appropriate uncertainty words to indicate the "Confidence provided at the end of "my knowledge" and use 'Internal Knowledge' to mark the source of the knowledge.
5. When citing the provided documents, you should select a fine-grained span from the documents and ensure the span is credible and less redundant. Use Roman numerals to mark the document and use Arabic numerals to mark spans. Use 'Document I' to refer to the first document, and so on.
6. Cite spans using Arabic numerals like [1]. Do not use Roman numerals to cite spans.
7. When using "my knowledge," you should generate a more credible and less redundant version of the knowledge, use Arabic numerals to mark the spans, and output the provided confidence in the last.
8. If none of "my knowledge" is available, admit it honestly and say that it is because of your limited capabilities.
9. If none of the documents and "my knowledge" is relevant to the question, you should still output the steps and an empty reference and then generate an abstention response: "I don't have sufficient knowledge to answer the question, and there is no relevant information in the provided documents to answer the question" with an empty reference.

Here is an example:
<example>


You have to follow the instructions to generate the full answer for the question below:

Question: <question>

`Documents`:
<docs>

`my knowledge`: 
<internal_knowledge>

Golden Answer: <golden_answer>

Output:
   \end{CustomVerbatim}
       \end{mdframed}
    \caption{Prompt for Rational Attribution and Elaboration generation}
    \label{fig:gen prompt}

\end{figure*}

\end{document}